\documentclass{article}
\usepackage{microtype}
\usepackage{graphicx}
\usepackage{subfigure}
\usepackage{booktabs}

\usepackage{hyperref}

\usepackage[accepted]{icml2020}

\usepackage{amsmath}
\usepackage{amssymb}
\usepackage{xcolor}
\usepackage{soul}
\usepackage[nolist,nohyperlinks]{acronym}

\begin{acronym}[fancy]
\acro{fancy}[GS-CIOC]{General-Sum Multi-Agent Continuous Inverse Optimal Control}
\end{acronym}

\usepackage[framemethod=tikz]{mdframed}
\usepackage{makecell}

\definecolor{mycolor}{rgb}{0.122, 0.435, 0.698}

\newmdenv[innerlinewidth=0.5pt, roundcorner=4pt,linecolor=mycolor,innerleftmargin=6pt,
innerrightmargin=6pt,innertopmargin=6pt,innerbottommargin=6pt]{mybox}

\usepackage{hyperref}

\icmltitlerunning{Diversity in Action}

\begin{document}
\twocolumn[
\icmltitle{Diversity in Action:\\
General-Sum Multi-Agent Continuous Inverse Optimal Control}

\icmlsetsymbol{equal}{*}

\begin{icmlauthorlist}
\icmlauthor{Christian Muench}{mer,d1}
\icmlauthor{Frans A. Oliehoek}{d2}
\icmlauthor{Dariu M. Gavrila}{d1}
\end{icmlauthorlist}

\icmlaffiliation{mer}{Environment Perception Group, Mercedes-Benz AG, Germany}
\icmlaffiliation{d1}{Intelligent Vehicles Group, TU Delft, The Netherlands}
\icmlaffiliation{d2}{Interactive Intelligence Group, TU Delft, The Netherlands}

\icmlcorrespondingauthor{Christian Muench}{c.muench@tudelft.nl}
\icmlkeywords{Machine Learning, ICML}

\vskip 0.3in
]

\printAffiliationsAndNotice{}

%%%%%%%%%%%%%%%%%%%%%%%%%%%%%%%%%%%%%%%%%%%%%%%%%%%%%%%%%%%%%%%%%%%%%%%%%%%%%%%%

\begin{abstract}
Traffic scenarios are inherently interactive. Multiple decision-makers predict the actions of others and choose strategies that maximize their rewards. We view these interactions from the perspective of game theory which introduces various challenges. Humans are not entirely rational, their rewards need to be inferred from real-world data, and any prediction algorithm needs to be real-time capable so that we can use it in an autonomous vehicle (AV). In this work, we present a game-theoretic method that addresses all of the points above.
Compared to many existing methods used for AVs, our approach does 1) not require perfect communication, and 2) allows for individual rewards per agent. Our experiments demonstrate that these more realistic assumptions lead to qualitatively and quantitatively different reward inference and prediction of future actions that match better with expected real-world behaviour.
\end{abstract}

\section{INTRODUCTION}
One of the most pressing problems that need to be solved to realize autonomous
driving in urban environments is the interaction with vulnerable road users such
as pedestrians and cyclists. Merely extrapolating the current velocity vector of
a pedestrian fails to account for behavioural changes caused by interactions
with traffic participants and the environmental layout (traffic lights, zebra
crossings, obstacles). 
Existing work \cite{Kretzschmar2016, Pfeiffer_2016_IROS, Sadigh2018, Schwarting2019, Ma} in the domain of intelligent vehicles may describe these kinds of
social interactions in terms of Markov Decision Processes (MDP), i.e. a framework
where each agent (human) is maximizing a reward (avoid collision, be
considerate). Three aspects are crucial when using MDPs to model human
interactions. One is inferring the correct reward function from real-world data,
another is to account for bounded rationality, and finally, we want real-time
predictions given a reward function.

\textbf{Reward Inference:}
One way of inferring rewards from multi-agent data is to fix the observed
actions of the other agents, i.e. decoupling the actions from the other agents
from one's own \cite{Schwarting2019, Sadigh2018}. Treating
other agents as dynamic obstacles reduces the problem to that of a single agent.
Another approach is to assume that all agents are controlled by one "brain"
\cite{Kretzschmar2016, Pfeiffer_2016_IROS}. Therefore, communication
between agents is instant, and they may coordinate their actions perfectly.
Additionally, all agents optimize the same (cooperative) reward function.  
While this assumption simplifies computations, it represents an overly
altruistic view of the world.

\textbf{Bounded Rationality:}
Humans are boundedly rational \cite{Wright2010}. Instead of executing an action
that leads to the highest expected reward, humans may choose another action that
rewards less. In other words, the probability of selecting an action increases
with the expected reward of that action. 
Humans do not follow deterministic paths in
traffic scenarios. Instead, they show natural variation in their decision making. 
\cite{Schwarting2019, Fridovich-Keil2019, Sadigh2018} argue from
the deterministic perspective that agents maximize their expected reward as much
as possible. We aim for a prediction algorithm that takes bounded
rationality into account.  

Another term that relates to the same concept is 
probability matching. \cite{Eysenbach2019} discusses probability matching in
maximum-entropy reinforcement learning and the connection to human
and animal decision making. Also, we may refer to the behaviour of an agent as
sub-optimal, meaning the same concept.

\textbf{Operational Constraints:}
A prediction algorithm should be capable of running efficiently (preferably in
real-time) and work in continuous action and state spaces.
\cite{Fridovich-Keil2019} demonstrate a multi-agent algorithm that is real-time
capable. The underlying reason for the efficiency is that the algorithm belongs
to the family of linear quadratic regulator (LQR) methods.

\textbf{Contributions:}
In this work, we address the challenges outlined above. We present a novel
algorithm that

\begin{itemize}
\item learns the reward functions of a diverse set of agents. It does not
assume the other agents as dynamic obstacles, and it does not assume instant
communication between agents.
\item accounts for variation in the decisions of an agent because it belongs
to the family of maximum-entropy algorithms.
\item can be adapted for prediction tasks, with the potential of running in
real-time as it is related to LQR methods.
\end{itemize}
In particular,
\begin{itemize}
    \item we extend the continuous inverse optimal control (CIOC)
    \cite{Levine2012} algorithm to the general-sum two-agent
    setting
    \item we verify the algorithm on simulations and show its usefulness. In
    particular, our algorithm allows us to choose different reward functions for
    each agent.
Additionally, we observe a significant difference in the deduced reward and predictions
when not assuming instant communication.
\end{itemize}

%%%%%%%%%%%%%%%%%%%%%%%%%%%%%%%%%%%%%%%%%%%%%%%%%%%%%%%%%%%%%%%%%%%%%%%%%%%%%%%%
\section{Related Work}
We focus on work that deals with traffic interactions involving pedestrians, or
that could be easily applied to such interactions. For a comprehensive overview
of the literature on predicting the intentions of pedestrians see
\cite{Rudenko2019}.
We focus on the multi-agent generalizations of Markov Decision Processes (MDPs)
where the agent (e.g. pedestrian) executes actions (step to the right) to
maximize a reward (avoid collision). 
A useful extension of the standard MDP framework to the maximum-entropy
framework can describe boundedly rational behaviour of humans more closely (see e.g.
\cite{Kitani}). 

In the following sub-sections, we outline other work close to ours, considering
(real-time) multi-agent games in continuous action and state spaces that
model interactions of robots (or humans) with humans in traffic like scenarios.

\textbf{Multi-Agent Games:}
\cite{Sadigh2018, Schwarting2019} consider so-called Stackelberg games
between cars in continuous state and action spaces where the agents take turns
and communicate their actions to the other agents before executing them.
This simplifies the computational complexity of the problem significantly since
it is sequential and deterministic in nature. Both publications infer the reward
functions from real-world data using CIOC \cite{Levine2012}. Though, other
agents are reduced to dynamic obstacles simplifying the reward inference (i.e.
non reacting).  

 The approaches of \cite{Schwarting2019a, Fridovich-Keil2019a, Fridovich-Keil2019} are inspired by the iLQG
 algorithm and the well-established solutions to linear quadratic games (see for
 example \cite{1982239}). They can deal with non-linear dynamics and non-linear
 cost functions in multi-agent dynamic games. In contrast to our work,
 \cite{Fridovich-Keil2019a, Fridovich-Keil2019, Schwarting2019a}
 do not consider boundedly rational agents and reward inference.

\textbf{Bounded Rationality:}
The notion of bounded rationality has a long tradition in both artificial intelligence \cite{simon1955behavioral, russell1997rationality, Zilberstein2011metareasoning, rubinstein1986finite, halpern2014decision}
and (behavioral) game theory \cite{McKelvey1995, rubinstein1998modeling, Wright2010}. In particular, we will use a model of bounded rationality that is very close to the quantal response equilibrium \cite{McKelvey1995}. Our contribution here can be seen as the ability to approximate such a QRE, in continuous games, while also inferring the rewards from data.

\textbf{Maximum Entropy Inverse Reinforcement Learning:} 
Also, in the context of RL, people have considered notions of boundedly rational agents.
\cite{Ziebart2008} introduced maximum-entropy inverse reinforcement learning (MaxEntIRL) to obtain the rewards from agents that act boundedly rational. 
The most daunting task in MaxEntIRL for high dimensional continuous action and
state spaces (i.e. no dynamic programming) is the derivation of the partition
function $Z=\int p(\tau)\exp(r(\tau))$. \cite{Kuderer2013} approximate the
distribution over trajectories with weighted sums of delta-functions
representing the observed data points, optimizing the data likelihood by
gradient ascent. A more advanced algorithm - which we will use in this paper -
is the use of the Laplace approximation (second-order Taylor expansion) around
the observed data points that models the curvature of the reward function
\cite{Levine2012, Dragan2013}.
Another possibility is a sampling-based approximation by, e.g. Monte Carlo
methods \cite{Kretzschmar2016, Pfeiffer_2016_IROS, Xu2019a}. In general, this corresponds to solving the full reinforcement learning
problem in an inner loop of the inverse reinforcement learning algorithm
\cite{Finn2016}. 

\section{BACKGROUND}
We are considering a two-agent stochastic game with shared states $x_t$, agent
specific actions $u_{it}$, $u_{jt}$ (i,j - agent index), agent specific rewards $r_i(x_t,
u_{it}, u_{jt})$ and stochastic transitions $p(x_t |u_{it}, u_{jt}, x_{t-1})$ to
the $x_t$ given the actions $u_{kt}$ and state $x_{t-1}$. In general, both the transitions
and the reward function depend on the actions of both agents\footnote{A fully
cooperative reward is an example where agent i may receive a reward for an
action that agent j executes.}.

Also, the rewards of the agents are discounted with a discount factor $\gamma$.
In the following, we give an overview of the most important formulas for the
single-agent case. These will translate to the two-agent setting naturally.

Given a state $x_{t-1}$ at time-step $t-1$ and an action $u_t$, an agent will transition to the next
state according to the stochastic environment transitions $p(x_t | x_{t-1},
u_t)$.\footnote{We follow the notation by \cite{Levine2012} in which the
action is indexed with the stage \emph{to which} it takes us.}
The agent will also receive a reward $r(x_t, u_t)$ depending on the action $u_t$
and the state $x_t$ that the environment (including the agent) transitions to.
In this work, we assume that an agent does not act fully rational and chooses sub-optimal actions. A natural description of this type of bounded rationality is the
maximum-entropy (MaxEnt) framework \cite{Ziebart2008} which can be used to describe the sub-optimal decisions of humans (e.g. \cite{Kitani}). In the MaxEnt framework, a trajectory is sampled from a probability
distribution given by
\begin{equation}
p(\tau)=\Pi_t p(x_{t}|x_{t-1}, u_{t})\exp(r(x_{t}, u_{t}))\label{eq:distribution}
\end{equation}
with the trajectory $\tau$ corresponding to the sequence of actions $u_{t}$ and
states $x_{t}$ over multiple time-steps $t$.
The policy $\pi(u_{t}|x_{t-1})$ of an agent, i.e. the conditional pdf that
describes the most likely actions $u_{t}$ that an agent takes given its current
state $x_{t-1}$ depends on the Q-function
\begin{equation}
    \pi(u_t|x_{t-1}) = \frac{\exp(Q(x_{t-1}, u_{t}))}{\int \exp(Q(x_{t-1},
    u_{t}') du_t')}\label{eq:policy}
\end{equation}
The Q-function may be derived by performing dynamic
programming, iterating over the soft-Bellman equation until convergence.
\begin{multline}
    Q(x_{t-1}, u_{t}) =  \int p(x_t' | u_{t}, x_{t-1}) \Big(r(x_{t}', u_{t}) 
    +\\ \gamma \log \int \exp(Q(x_{t}', u_{t+1}')) du'_{t+1}\Big)dx_t' \label{eq:softbellmansingle}
\end{multline}
The second term inside the first integral is the value function.
\begin{equation}
    V(x_{t}) = \log \int \exp(Q(x_{t}, u'_{t+1})) du'_{t+1}\label{eq:value}
\end{equation}
The reason we refer to (\ref{eq:softbellmansingle}) as the soft-Bellman equation is the
soft-maximization operator $\log \int \exp$. In contrast, the standard Bellman equation
employs the "hard" maximization operator $\max$. A connection can be established
by scaling the reward function and considering the limit of $\lim_{\alpha
\rightarrow +\infty}(\alpha r)$ which
will recover the "hard" maximization in the soft-Bellman equation and a
policy that satisfies the standard Bellman equation given the unscaled reward
function.
Please refer to the excellent tutorial on maximum-entropy reinforcement learning
and its connection to probabilistic inference \cite{Levine2018} for a thorough
derivation of the soft-Bellman equation and its connection to
(\ref{eq:distribution}).

In the following sections, we will only consider finite horizon problems, i.e. each agent will collect rewards for a limited amount of time. Additionally, we set the discount factor to $\gamma = 1$.

\section{General-Sum Multi-Agent Continuous Inverse Optimal Control}
We extend CIOC to the two-agent\footnote{Extension to N agents is discussed in supplementary material.} setting where each agent may
receive a different reward. A major difference to the derivation of CIOC is the
environment transitions that are not deterministic anymore. We assume that the
other agent is part of the environment and acts according to a stochastic policy. The iterative nature of the algorithm bears a resemblance to \cite{nair2003taming}, though, we update the policies in parallel (not alternating) while also being able to handle continuous states and actions.
A major advantage of CIOC and its extension is the relative ease of inferring
the reward parameters from demonstrations. We can backpropagate the gradients directly through the policy, eliminating the need to run a complex deep
reinforcement learning algorithm every time we update the reward parameters.

We will present three algorithms that are interconnected. \ac{fancy} (algorithm \ref{al:gscioc}) that returns policies for quadratic rewards and linear environment transitions. For non-quadratic rewards and non-linear transitions, Iterative \ac{fancy} can be used (algorithm \ref{al:itgscioc}) to obtain locally optimal policies. It uses \ac{fancy} as a sub-routine. Finally, the reward inference algorithm \ref{al:rewardinfer} uses \ac{fancy} for obtaining local policy approximations around observed real-world data for any type of reward functions and transitions.

\subsection*{Two-Agent Soft-Bellman Equation}
We assume that all agents choose their
actions in accordance with (\ref{eq:distribution}) which directly raises the
question how they deal with the presence of other agents. Here, the other agent
is part of the environment, similar to the multi-agent setting in interacting
Partially Observable Markov Decision Processes (POMDPs) as described by
\cite{Gmytrasiewicz2005}. 
Let $u_{it}$, $u_{jt}$ be the actions of agent i and j respectively. We assume
the environment transitions to be deterministic, i.e. $p(x_t|x_{t-1}, u_{it},
u_{jt})$ is a
deterministic function. The reward function of agent i may depend on the action
of agent j, i.e. $r_i(x_t, u_{it}, u_{jt})$, though $u_{jt}$ is not known to
agent i at $t-1$. Therefore, for argument's sake, we define a new state variable
$\tilde x_t = [x_t, u_{jt}]$ and the corresponding stochastic environment transitions
$p(\tilde x_t|\tilde x_{t-1}, u_{it}) := p(x_t, u_{jt}|x_{t-1}, u_{jt-1}, u_{it})$. 
The soft-Bellman equation for agent i is
as follows
\begin{multline}
    Q_i(\tilde x_{t-1}, u_{it}) = 
    \int p(\tilde x_t|\tilde x_{t-1}, u_{it}) \Big(r_i(\tilde x_{t},
    u_{it}) \\+ \gamma \log \int \exp(Q_i(\tilde x_{t}, u_{it+1}))
    du_{it+1}\Big)d\tilde x_{t} %\label{eq:softbellman}
\end{multline}
The agent decides on its action based on $x_{t}$, not $\tilde x_t$. Thus, we can
drop the $\tilde x_t$ dependency in the Q-function. Also, we can expand the
environment transitions $p(\tilde x_t|\tilde x_{t-1}, u_{it}) = p(x_t|x_{t-1},
u_{it}, u_{jt})\pi_j(u_{jt}|x_{t-1})$, with $\pi_j$ being the policy of agent j.
The soft-Bellman equation can now be reformulated as
\begin{multline}
    Q_i(x_{t-1}, u_{it}) = \\
    \int p(x_t|x_{t-1}, u_{it}, u_{jt})\pi_{j}(u_{jt}|x_{t-1})
    \Big(r_i(x_{t},
    u_{it}, u_{jt}) \\+ \gamma \log \int \exp(Q_i(x_{t}, u_{it+1}))
    du_{it+1}\Big)du_{jt}dx_t \label{eq:softbellman}
\end{multline}

Solving the Bellman-equation for high-dimensional continuous state and action
spaces is in general intractable. Tackling this problem employs many researchers
in the fields of reinforcement learning and optimal control. Here, we make use of
the so-called Laplace approximation that deals with the difficulty of
calculating the partition function $\int \Pi_t p(x_t|x_{t-1}, u_t)\exp(r(x_t,
u_t))du_tdx_t$ by approximating the reward function with a second-order Taylor
expansion. \cite{Levine2012} developed the Continuous Inverse Optimal Control
(CIOC) algorithm based on this approximation. Though, their approach is limited
to the single-agent case with deterministic environment transitions. We will
show how to extend their algorithm to the two-agent setting.
\subsection*{Value Recursion Formulas}
\begin{algorithm}[tb]
   \caption{\ac{fancy}}
   %\label{alg:example}
\begin{algorithmic}
   %\STATE {\bfseries Input:} reference trajectories \\$\tau_k = \{(x_{t}, u_{kt}),\quad t
   %\in \{1, ..., T\}\}$, $k \in \{i, j\}$
   \STATE {\bfseries Input:} Reference trajectories $\tau_k$, $k\in\{i,j\}$
   \STATE Taylor expansion (\ref{eq:taylorreward}) of rewards along $\tau_k$
   \STATE Initialize $V_k(x_T) \leftarrow 0$
   \FOR{$t\leftarrow T$ {\bfseries to} $1$}
   \STATE \COMMENT{Update Gaussian policy:}
   \STATE $\mu_{kt}\leftarrow$ Solve (\ref{eq:meanaxb}) for mean action   
   \STATE $\tilde M_{(kk)t}\leftarrow$ Determine precision matrix (\ref{eq:precision})
   \STATE \COMMENT{Recompute value function, given updated policy:} 
   \IF{$t > 1$}
   \STATE $V_k(x_{t-1})\leftarrow$ Value recursion (\ref{eq:valuerecursionmain}) -
   (\ref{eq:valuerecursionmainend})

   \ENDIF
   \ENDFOR
   \STATE {\bfseries Return:} Policies $\pi_i$ and $\pi_j$
   \label{al:gscioc}
\end{algorithmic}
\end{algorithm}

The procedure that we obtain in this section is illustrated in algorithm
\ref{al:gscioc}. We take a reference trajectory $\tau_k$ - a sequence of states $x_t$ and
actions $u_{kt}$ - of each agent $k\in(i, j)$ and approximate the reward
function $r_k$ close to the reference trajectory. This will allow us to derive a
local policy approximation $\pi_k$ - an approximation that works best if the
agent stays close to the reference trajectory - by working our way from the end
of the reference trajectory to the beginning calculating the value function
$V_k(x_t)$ for each time-step. In other words, the formulas are recursive in
nature. Where does the reference trajectory come from? It may be a randomly
chosen state and action sequence, or it may represent actual observation data.

We sketch the derivation of algorithm \ref{al:gscioc} starting at the final
time-step (the horizon) of the reference trajectories $\tau_i$, $\tau_j$. Here the Q-function of agent i and by extension the policy $\pi_i$ can be calculated as follows
\begin{multline}
    Q_{i}(\bar x_{T-1}, \bar u_{iT}) = \int p(\bar x_T|\bar
    x_{T-1},\bar u_{iT},\bar u_{jT})
    \\\pi_{j}(\bar u_{jT}|\bar x_{T-1}) r_i(\bar x_{T}
    , \bar u_{iT}, \bar u_{jT})d\bar u_{jT}d\bar x_{T}\label{eq:lastQ}
\end{multline}
Even though $p(\bar x_T|\bar x_{T-1},\bar u_{iT},\bar u_{jT})$ is deterministic,
the integral is intractable in general. We circumvent this problem by
approximating the reward function $r_i$ using a second-order Taylor expansion around
the fixed reference trajectories\footnote{Here, we stay close to the
single-agent LQR derivation in \cite{Levine2012} where actions and states
separate in the reward function $r(x_t, u_t) = g(x_t)+f(u_t)$. This is also the
structure that we assume in the experiments in the experimental section. For a derivation
that considers more general reward functions, please refer to the supplementary
material.}. 
\begin{multline}
    r_i(\bar x_{t}, \bar u_{it}, \bar u_{jt}) \approx r_{it} + \bar
    x_{jt}^T\Big(\hat H_{(ji)t}\bar x_{it}\Big) + \bar u^T_{jt}\Big(\tilde
    H_{(ji)t}\bar u_{it} \Big) \\+\sum_{k=1}^2 \Bigg[\bar
    x_{kt}^T\Big(\frac{1}{2}\hat H_{(kk)t}\Big)\bar x_{kt} + \bar u^T_{kt}
    \Big(\frac{1}{2}\tilde H_{(kk)t}\Big)\bar u_{kt} + \\\bar u_{kt}^T\tilde
    g_{kt} + \bar x_{kt}^T\hat g_{kt}\Bigg] \label{eq:taylorreward}
\end{multline}
$H$ and $g$ refer to the Hessians and gradients w.r.t the states and actions of
both agents. 
\begin{align}
\hat H_{(nm)t} &= \frac{\partial^2 r_i}{\partial \bar x_{nt} \partial \bar
x_{mt}}, \qquad
\tilde H_{(nm)t} = \frac{\partial^2 r_i}{\partial \bar u_{nt} \partial \bar
u_{mt}}\\
\hat g_{nt} & = \frac{\partial r_i}{\partial \bar x_{nt}}, \qquad
\tilde g_{nt} = \frac{\partial r_i}{\partial \bar u_{nt}}
\end{align}
The reward function is now quadratic with
\begin{equation}
    \bar x_{t} = x_{t} - x_{t}^*, \qquad
    \bar u_{kt} = u_{kt} - u_{kt}^*
\end{equation}
where $x^*$ and $u^*$ refer to the fixed reference trajectories. 
The state $\bar x_t = [\bar x_{it}, \bar x_{jt}]$ is split into the agent
specific sub-states which are directly controlled by each agent.

We need one additional approximation to solve the integral in (\ref{eq:lastQ}).
Namely, we linearize the dynamics
\begin{align}
    \bar x_{kt} & = A_{kt} \bar x_{kt-1} + B_{kt} \bar u_{kt}\\
    A_{kt} & = \frac{\partial \bar x_{kt}}{\partial \bar x_{kt-1}}, \qquad
    B_{kt} = \frac{\partial \bar x_{kt}}{\partial \bar u_{kt-1}}
\end{align}
Applying the linearization and the quadratic approximation of the reward
function turns (\ref{eq:lastQ}) into a tractable integral. In particular,
$Q_{iT}$ is quadratic in the actions $\bar u_{iT}$. Therefore, the policy of
agent i is a Gaussian policy. The same is true for agent j since we apply the
same approximations, i.e. $\pi_j$ is a Gaussian policy. The mean of that policy
is (general result)
\begin{equation}
    \mu_{it} = -\tilde M^{-1}_{(ii)t}\Big(\tilde g_{it} + \tilde H_{(ij)t}\mu_{jt}
    + B^T_{it}\hat q_{ijt} + B^T_{it}\hat Q_{(ii)t}A_{it}\bar x_{it-1}\Big)
    \label{eq:meanmain}
\end{equation}
$\tilde M_{(ii)t}$ is the precision matrix of the Gaussian policy of agent i.
\begin{equation}
    \tilde M_{(nm)t} = B_{nt}^T\hat Q_{(nm)t}B_{mt}+\tilde
    H_{(nm)t}\label{eq:precision}
\end{equation}
Additional definitions of symbols used in the equations above
\begin{equation}
    \hat q_{it} = \hat g_{it} + \hat v_{it}, \qquad
    \hat Q_{(nm)t} = \hat H_{(nm)t} + \hat V_{(nm)t}
\end{equation}
(\ref{eq:meanmain}) is a system of linear equations in $\mu_{it}$ and $\mu_{jt}$ of
the form $Ax=b$.
\begin{align}
    &\Big[\delta_{ij}^{(i)}\Big(\tilde M_{(jj)t}^{-1}\Big)^{(j)}\delta_{ji}^{(j)} -
    \Big(\tilde M_{(ii)t}\Big)^{(i)}\Big]\mu_{it} =\nonumber\\ &\alpha_i^{(i)} -
    \delta_{ij}^{(i)}\Big(\tilde M_{(jj)t}^{-1}\Big)^{(j)}\alpha_j^{(j)} +
    \nonumber\\ &\Big(\beta_i^{(i)} - \delta_{ij}^{(i)}\Big(\tilde
    M_{(jj)t}^{-1}\Big)^{(j)}\gamma_{ji}^{(j)}\Big)\bar x_{it-1} +\nonumber\\
    &\Big(\gamma_{ij}^{(i)} - \delta_{ij}^{(i)}\Big(\tilde
    M_{(jj)t}^{-1}\Big)^{(j)}\beta_{j}^{(j)}\Big)\bar x_{jt-1}\label{eq:meanaxb}
\end{align}
with
\begin{align}
    \alpha_i^{(i)} = & \Big[\tilde g_{it} + B_{it}^T\hat q_{it}\Big], \qquad
    \beta_i^{(i)} = B_{it}^T\hat Q_{(ii)t}A_{it}\\
    \gamma_{ij}^{(i)} = & B_{it}^T\hat Q_{(ij)t}A_{jt}, \qquad
    \delta_{ij}^{(i)} = \tilde M_{(ij)t}
\end{align}
The $^{(i)}$ index indicates which agent the derivatives/ value functions refer to. We do not only recover the mean of agent j but also that of agent i.

At last, we calculate the value function using equation (\ref{eq:value}). The
Q-function is a quadratic polynomial in the states and actions. Therefore, the
integral is tractable, and we end up with a value function with the following
structure
\begin{multline}
    V_{i}(\bar x_{T}) =
    \bar x^T_{jT} \Big(\hat V_{(ji)T}\bar x_{iT}\Big) +\\ \sum_{k=1}^2\left[\bar
    x^T_{kT} \Big(\frac{1}{2}\hat V_{(kk)T}\Big)\bar x_{kT} + \bar x^T_{kT}\hat
    v_{kT}\right] + const\label{eq:valuequad}
\end{multline}
How do we obtain the $\hat V$ and $\hat v$ matrices? We collect all the
terms that are quadratic or linear in $\bar x_{kt}$. The general result is given
in equation (\ref{eq:valuerecursionmain}) onward.

Now, that we know $V_{iT}$, we apply the same procedure to the previous time
step, i.e. the calculations are repetitive. In short, we derive the Q-function
of the next time-step ($T-1$).
\begin{multline}
    Q_{i}(\bar x_{T-2}, \bar u_{iT-1}) = \\\int p(\bar x_{T-1}|\bar
    x_{T-2},\bar u_{iT-1},\bar u_{jT-1}) 
    \pi_{j}(\bar u_{jT-1}|\bar x_{T-2}) 
    \Big(\\r(\bar x_{T-1}, \bar u_{iT-1}, \bar u_{jT-1}) + V_{i}(
    \bar x_{T})\Big)d\bar u_{jT-1}d\bar x_{T-1}
\end{multline}
Given the reward approximation, the linearization of the dynamics and the
quadratic state dependency of the value function $V_{iT}$ in equation
(\ref{eq:valuequad}) this integral is tractable again. Working our way along the
full trajectory, we obtain recursive formulas for the value functions (meaning
the $\hat V$ matrices and $\hat v$ vectors in (\ref{eq:valuequad})). Due to
space constraints, we will state the results and refer the interested
reader to the supplementary material.
\begin{align}
    \hat V_{(ii)t-1} 
    &= \Pi_{jt}^T \Big[\tilde M_{(jj)t} - \tilde M_{(ji)t}\tilde M_{(ii)t}^{-1}\tilde M_{(ij)t}]\Pi_{jt} \nonumber\\ 
    +&\Big[A_{it}^T\hat Q_{(ij)t}B_{jt} - A_{it}^T\hat Q_{(ii)t}B_{it}\tilde M_{(ii)t}^{-1}\tilde M_{(ij)t}\Big]\Pi_{jt}\nonumber\\
    +&\Pi_{jt}^T\Big[A_{it}^T\hat Q_{(ij)t}B_{jt} - A_{it}^T\hat Q_{(ii)t}B_{it}\tilde M_{(ii)t}^{-1}\tilde M_{(ij)t}\Big]^T\nonumber\\
    \textcolor{blue}{+}&\textcolor{blue}{A^T_{it}\hat Q_{(ii)t}A_{it}}
    \textcolor{blue}{-A_{it}^T\hat Q_{(ii)t}^TB_{it}\tilde M_{(ii)t}^{-1}B_{it}^T\hat
    Q_{(ii)t}A_{it}}\label{eq:valuerecursionmain}
\end{align}
\begin{align}
    \hat V_{(jj)t-1} &= \Omega^T_{jt}\Big[\tilde M_{(jj)t} - \tilde M_{(ji)t}\tilde M_{(ii)t}^{-1}\tilde M_{(ij)t}\Big]\Omega_{jt} \nonumber\\
    +& \Big[A^T_{jt}\hat Q_{(jj)t}B_{jt} - A^T_{jt}\hat Q_{(ji)t}B_{it}\tilde M_{(ii)t}^{-1}\tilde M_{(ij)t}\Big]\Omega_{jt}\nonumber\\
    +& \Omega_{jt}^T\Big[A^T_{jt}\hat Q_{(jj)t}B_{jt} - A^T_{jt}\hat Q_{(ji)t}B_{it}\tilde M_{(ii)t}^{-1}\tilde M_{(ij)t}\Big]^T \nonumber\\
    +& A_{jt}^T\hat Q_{(jj)t}A_{jt} - A_{jt}^T \hat Q_{(ji)t}B_{it}\tilde M_{(ii)t}^{-1}B_{it}^T\hat Q_{(ij)t}A_{jt}
\end{align}
\begin{align}
    \hat V_{(ji)t-1} &= \Omega_{jt}^T\Big[\tilde M_{(jj)t} - \tilde M_{(ji)t}\tilde M_{(ii)t}^{-1}\tilde M_{(ij)t}\Big]\Pi_{jt} \nonumber\\
    +& \Omega_{jt}^T\Big[A_{it}^T\hat Q_{(ij)t}B_{jt} - A_{it}^T\hat Q_{(ii)t}B_{it}\tilde M_{(ii)t}^{-1}\tilde M_{(ij)t}\Big]^T \nonumber\\
    +& \Big[A_{jt}^T\hat Q_{(jj)t}B_{jt} - A_{jt}^T\hat Q_{(ji)t}B_{it}\tilde M_{(ii)t}^{-1}\tilde M_{(ij)t}\Big]\Pi_{jt} \nonumber\\
    +& A_{jt}^T\hat Q_{(ji)t}A_{it} - A_{jt}^T\hat Q_{(ji)t}B_{it}\tilde M_{(ii)t}^{-1}B_{it}^T\hat Q_{(ii)t}A_{it}
\end{align}
\begin{align}
    \hat v_{jt-1} &= \Omega_{jt}^T\Big[\tilde M_{(jj)t} - \tilde M_{(ji)t}\tilde M_{(ii)t}^{-1}\tilde M_{(ij)t}\Big]\nu_{jt}\nonumber\\
    +&\Big[A^T_{jt}\hat Q_{(jj)t}B_{jt} - A^T_{jt}\hat Q_{(ji)t}B_{it}\tilde M_{(ii)t}^{-1}\tilde M_{(ij)t}\Big]\nu_{jt}\nonumber\\
    +& \Omega_{jt}^T\Big(\tilde g_{jt} + B_{jt}^T\hat q_{jt} - \tilde M_{(ji)t}\tilde M_{(ii)t}^{-1}\Big[\tilde g_{it} + B_{it}^T\hat q_{it}\Big]\Big)\nonumber\\
    +& A_{jt}^T\hat q_{jt} - A_{jt}^T\hat Q_{(ji)t}B_{it}\tilde M^{-1}_{(ii)t}[\tilde g_{it} + B^T_{it}\hat q_{it}]
\end{align}
\begin{align}
    \hat v_{it-1} &= \Pi_{jt}^T\Big[\tilde M_{(jj)t} - \tilde M_{(ji)t}\tilde M_{(ii)t}^{-1}\tilde M_{(ij)t}\Big]\nu_{jt} \nonumber\\
    +&\Big[A^T_{it}\hat Q_{(ij)t}B_{jt} - A^T_{it}\hat Q_{(ii)t}B_{it}\tilde M_{(ii)t}^{-1}\tilde M_{(ij)t}\Big]\nu_{jt} \nonumber\\
    +&\Pi_{jt}^T\Big(\tilde g_{jt} + B_{jt}^T\hat q_{jt} - \tilde M_{(ji)t}\tilde M_{(ii)t}^{-1}\Big[\tilde g_{it} + B_{it}^T\hat q_{it}\Big]\Big)\nonumber\\
    \textcolor{blue}{+}& \textcolor{blue}{A^T_{it}\hat q_{it}}\textcolor{blue}{
    -A^T_{it}\hat Q_{(ii)t}B_{it}\tilde M^{-1}_{it}\Big[\tilde g_{it} +
    B^T_{it}\hat q_{it}\Big]}\label{eq:valuerecursionmainend}
\end{align}
The parts of the formulas that are highlighted in \textcolor{blue}{blue}
correspond to the single agent value recursion formulas from \cite{Levine2012}.
If the reward is a single agent reward function, the value recursion formulas
will reduce to the highlighted parts.

$\Pi$ and $\Omega$ refer to the reactive policy of agent j, i.e. they encode how
agent j will react to deviations in the position of agent i (or agent j itself). The
mean action of agent j is given by
\begin{equation}
    \mu_{jt} = \nu_{jt} + \Pi_{jt}\bar x_{it-1} + \Omega_{jt} \bar x_{jt-1}
\end{equation}
\subsection*{Iterative \ac{fancy}}
\begin{algorithm}[tb]
   \caption{Iterative \ac{fancy}}
   %\label{alg:example}
\begin{algorithmic}
   \STATE {\bfseries Input:} Initial trajectories $\tau_i$, $\tau_j$ learning rate $\eta$
   \REPEAT
   \STATE \textbf{$\pi_{i}$, $\pi_{j}\leftarrow$ GS-CIOC( $\tau_i$, $\tau_j$) } 
   \STATE \COMMENT{Update $\tau_i$, $\tau_j$} 
   \STATE Roll-out Gaussian policies with scaled mean $\eta \mu_{kt}$ and
   covariance = 0 (i.e. deterministic policy)
   \UNTIL{max iterations}
   \STATE {\bfseries Return:} Policies $\pi_i$ and $\pi_j$
   \label{al:itgscioc}
\end{algorithmic}
\end{algorithm}
We can use the algorithm described above to generate predictions (rollout policy), but this will only result in globally optimal solutions for quadratic rewards and linear (deterministic) environment transitions. In that case, the Taylor expansion around any given reference trajectory is exact everywhere.
If the rewards are not quadratic, the quadratic approximation of the reward and value function is, of course, inaccurate and holds only close to a given reference trajectory. Though,
we can execute \ac{fancy} iteratively to find local optima (with no convergence guarantee). We start with an initial trajectory and approximate the rewards and dynamics locally. We obtain a Gaussian policy by executing the value
recursion described above and update the trajectory by following the policy up
to a learning rate $\eta$. One possibility is to scale the mean of the policy
via $\eta\mu$. This way, we can control how much the update deviates from the
reference trajectory. This algorithm resembles the iLQG algorithm and has been
explored recently for the multi-agent setting in dynamic games by \cite{Schwarting2019a, Fridovich-Keil2019a}. We illustrate the procedure in algorithm \ref{al:itgscioc}.

In general, it is important to note that neither \cite{Schwarting2019a} nor
\cite{Fridovich-Keil2019a}, nor our approach recover Nash-equilibria (NE) since
the best responses
computed are approximations that work only locally
w.r.t. the current reference trajectory. The agents might not find a better
response which lies far away from the current reference trajectory. See also
\cite{Oliehoek2019} for a discussion on why local NE can be far from a NE.

\section{Recovering Reward Parameters}

Until now, we have discussed how to construct a local policy given a set of 
reference trajectories and a reward function (see algorithms \ref{al:gscioc} and
\ref{al:itgscioc}). However, since our goal is to use the predicted behaviours in
real-life traffic interactions, we need the capability to infer the parameters
$\theta$ of a reward function that captures human behaviour. 
For this purpose, we present an algorithm that can be run before deploying algorithm \ref{al:itgscioc}, to infer realistic rewards. Specifically, given observation
data of, e.g. a pedestrian interacting with a car, we can infer the reward parameters by maximizing the log-likelihood of the observed data
\begin{align} 
\theta^* &= \nonumber\\\arg& \max_{\theta}\frac{1}{|\tau|}\sum_{\tau, (u, x)\in
\tau}\ln p_\theta(x_{i0:T-1}, x_{j0:T-1}, u_{i1:T},
u_{j1:T})\nonumber\\ &=\arg \max_{\theta} \frac{1}{|\tau|}\sum_{\tau, (u, x)\in
\tau}\Bigg(\sum_{t}\ln \pi_{i, \theta}(u_{it}|x_{it-1}, x_{jt-1}) \nonumber\\ &+
\sum_{t}\ln \pi_{j, \theta}(u_{jt}|x_{it-1}, x_{jt-1})\Bigg)
\label{eq:irlobjective}
\end{align}
$\theta$ refers to the reward parameters and $|\tau|$ to the number of
trajectories $\tau$ in the data set. $\pi_{i, \theta}$ corresponds to the policy
of agent i (e.g. pedestrian) and is calculated with \ac{fancy}. Indeed, it is possible to
perform backpropagation through the entire \ac{fancy} algorithm. The procedure
is illustrated in algorithm \ref{al:rewardinfer}.
We optimize the objective (\ref{eq:irlobjective}) using gradient ascent.
Overall, the approach is similar to the single-agent reward inference of CIOC.

\begin{algorithm}[tb]
   \caption{Reward Inference}
   %\label{alg:example}
\begin{algorithmic}
   \STATE {\bfseries Input:} $\tau_i$, $\tau_j$ from data, initial
   $\theta$
   \REPEAT
   \STATE \textbf{$\pi_{i,\theta}$, $\pi_{j,\theta}\leftarrow$ GS-CIOC( $\tau_i$, $\tau_j$ ) } 
   \STATE Gradient ascent step on objective (\ref{eq:irlobjective}) 
   \UNTIL{max iterations or convergence of $\theta$}
   \STATE {\bfseries Return:} Reward parameters $\theta$
   \label{al:rewardinfer}
\end{algorithmic}
\end{algorithm}

\section{DIVERSITY IN ACTION}

In the following section, we demonstrate the capability of \ac{fancy} and the main
difference to a centralized multi-agent formulation of CIOC that we will call M-CIOC (for multi-agent). 
CIOC is originally a single-agent algorithm, though, we can transform it into a
multi-agent algorithm by assuming both agents as one (four-dimensional actions
instead of two). This is a typical approach in the literature used to model multi-agent interactions in a simplified way (see for example,
\cite{Kretzschmar2016, Pfeiffer_2016_IROS}).
Everything is implemented using JAX \cite{jax2018github}.
\begin{figure}[h]
\includegraphics[width=0.15\textwidth]{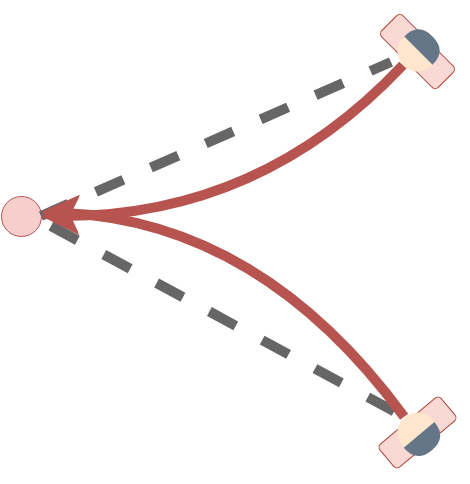}
\centering
\caption{Two agents (e.g. pedestrians) move towards a goal (red circle). The
dashed lines indicate the shortest path. Due to an interaction reward, the
agents are incentivized to walk towards each other.} \label{fig:lqr_overview} 
\end{figure} 

\begin{figure}[h]
\includegraphics[width=0.47\textwidth]{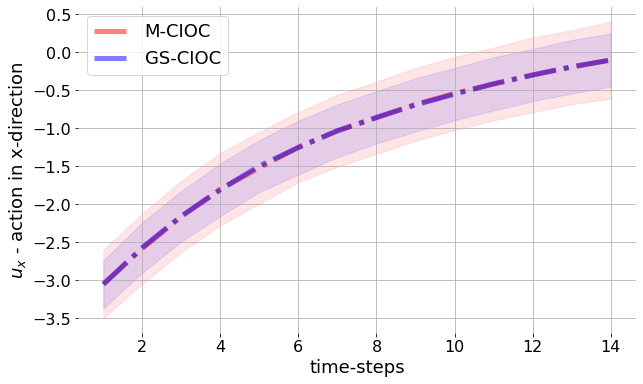}
\centering
\caption{We illustrate the actions of an agent for the scenario described in figure
\ref{fig:lqr_overview}. The reward setup is fully cooperative, i.e. both agents
receive the same reward. Both M-CIOC and \ac{fancy} can be applied to roll-out
action sequences. Given 2000 roll-outs, the mean actions are in agreement 
between the two algorithms. Though, the standard deviation (shaded region)
varies significantly. The reason for this is that agents can coordinate
perfectly in M-CIOC. Therefore, one agent can compensate for larger deviations of
the other agent from the mean and vice versa, leading to larger possible action amplitudes around the mean (they can minimize the interaction reward more effectively). \ac{fancy} assumes uncoordinated
execution of the action sequences, thus, exhibiting lower variance (interaction reward constricts movement around mean trajectory).\label{fig:ciocdec}}
\end{figure}

\textbf{Quadratic Rewards:}
The following setup is useful for understanding the validity of the
implementation of \ac{fancy} and one of the differences to  M-CIOC. We illustrate the scenario in figure \ref{fig:lqr_overview}:
Two agents move towards a goal and are pulled together by an interaction reward,
similar to people who belong together form a small group. We assume a
cooperative reward function of the form $r=-\alpha_1(x_1^2 + x_2^2)
-\alpha_2(u_1^2 + u_2^2) - \alpha_3(u_1 + u_2)^2$ (reward parameters, $\alpha_i
\ge 0$, $x$-states, $u$-actions), i.e. both agents maximize the same reward\footnote{They do so sub-optimally, which is why their actions show variance.}. The dynamics are linear with
$x_t = x_{t-1} + u_t$. We initialize one agent at $x=y=20$
and the other at $x=20, y=-20$. 
The reward function incentivizes both agents to
move towards $x=y=0$.  Though, they cannot do so in one step as non-zero actions
are penalized quadratically.  $-\alpha_3(u_1 + u_2)^2$ is special in that it induces an
interaction of the agents. In particular, the term is maximized if $u_1 = -u_2$.
We roll-out 2000 trajectories over $T=14$ time-steps using M-CIOC and \ac{fancy}.
Figure \ref{fig:ciocdec} depicts
the results. The mean solution of M-CIOC and \ac{fancy} is practically identical,
while the variances in the actions of each agent differ significantly. 

The reason for this is as follows. M-CIOC assumes agents
that can coordinate their actions perfectly. This is not the case for \ac{fancy}.
While the deviations from the mean trajectory are almost decorrelated for
\ac{fancy} with a correlation coefficient of $-0.1$ (correlation between agent 1
and 2), those of M-CIOC are highly correlated with a correlation coefficient of $-0.7$. The
variance resulting from M-CIOC algorithm is larger by a factor of
$1.9$ than that for \ac{fancy}. The reason is that if agent 1 chooses a
specific action $\Delta u_1$ ($\Delta u_1$, deviation from mean trajectory/
expected action), then agent 2 can choose the action $\Delta u_2 = -\Delta u_1$, canceling each other out in the interaction reward $-\alpha_3(u_1 + u_2)^2$. Through
coordination, the agents experience a wider range of possible actions in M-CIOC.
This is not necessarily a desirable property as we will demonstrate for the task
of inferring the reward function.  

To test how well the proposed algorithm can recover rewards from behaviour, we generate data trajectories based on a ground truth reward function.
Given 2000 roll-outs of \ac{fancy} we apply
algorithm \ref{al:rewardinfer}. 
We initialize the parameters at $\alpha=[1,1,1]$ and
perform gradient ascent with a learning rate of $10^{-3}$ until the objective converges. 
The result is $\alpha_1 = 0.2 (0.2)$, $\alpha_2 = 1.0 (1.0)$, $\alpha_3 = 3.0 (3.0)$. The values in the brackets indicate the true reward parameters. As we can see algorithm \ref{al:rewardinfer} recovered the rewards successfully.

What if we apply algorithm \ref{al:rewardinfer} to M-CIOC 2000 roll-outs?
Again, we initialize the parameters at $\alpha=[1,1,1]$ and perform gradient
ascent until convergence. We obtain: $\alpha_1 = 0.1 (0.2)$, $\alpha_2 = 0.5
(1.0)$, $\alpha_3 = 1.6 (3.0)$. The inferred reward is almost half of the
true reward, which matches the difference in observed variances of a factor of
$1.9$. Depending on the application, either a reward inference algorithm based on M-CIOC or \ac{fancy} will be more
accurate (assuming fully cooperative reward functions). If a centralized controller controls the agents, then M-CIOC should be preferred, though, for decentralized controllers \ac{fancy} is the better choice (i.e. algorithm \ref{al:rewardinfer}).  

We also tested the reward inference using
different reward parameters with $\alpha_1=0.4 (0.4)$, $\alpha_2 = 1.5 (1.5)$,
$\alpha_3 = 2.5 (2.5)$ for one agent and $\alpha_1=0.2 (0.2)$, $\alpha_2 = 1.0
(1.0)$, $\alpha_3 = 3.0 (3.0)$ for the other agent. Again, we were able to
reproduce the parameter values using algorithm \ref{al:rewardinfer} (2000 trajectories,
initialization at $\alpha=[1,1,1]$, \ac{fancy} for roll-out
and local policy approximation). In this case, M-CIOC cannot be used since it cannot handle different rewards per agent.

\begin{figure}[h]
\includegraphics[width=0.27\textwidth]{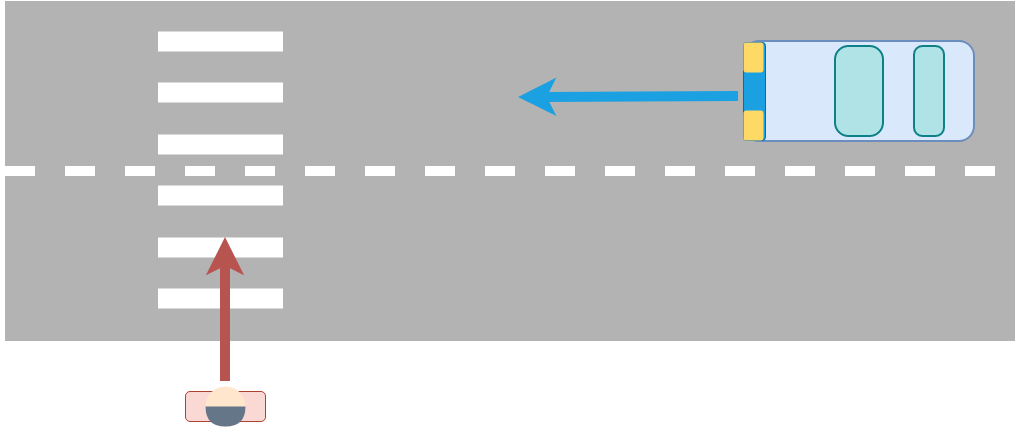}
\centering
\caption{A \textcolor{red}{pedestrian} approaches a zebra crossing. The
\textcolor{blue}{car} will yield, though
the pedestrian may be more or less socially inclined, speeding up so that the
car can go on earlier or continue to walk at her most comfortable pace.} \label{fig:zebra_overview} 
\end{figure} 

\begin{figure}[h]
\includegraphics[width=0.47\textwidth]{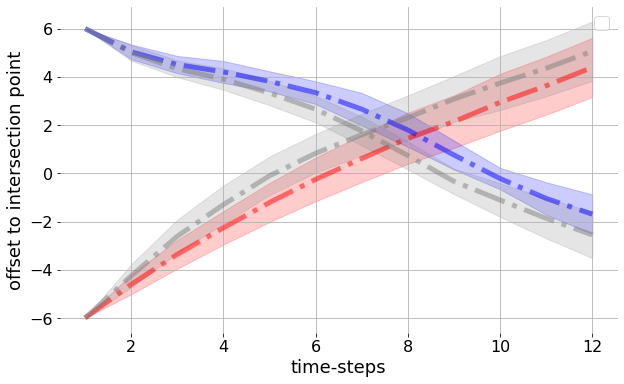}
\centering
\caption{The \textcolor{red}{pedestrian} crosses the zebra crossing with an
approaching \textcolor{blue}{car}
waiting until the pedestrian is on the other side (see also figure
\ref{fig:zebra_overview}). The y-axis indicates the
position relative to where the car and the pedestrian path intersect. The red
(blue) curve corresponds to a pedestrian (car) that only considers her own reward. The grey
curve corresponds to a pedestrian that values the progress of the car.
Therefore, the latter speeds up in the beginning to make room for the car. 
The shaded regions indicate one standard deviation around the mean trajectory.}
\label{fig:zebra} 
\end{figure}

\textbf{Traffic Interaction Scenario:}
A major advantage of \ac{fancy} is the ease of defining reward functions for
separate agents. 
Furthermore, in contrast to deterministic best responses, the maximum-entropy formulation is a form of reward-proportional error prediction, which was shown to capture human (boundedly-rational) behaviour well \cite{Wright2010}.
We demonstrate the interaction of
a vehicle and a pedestrian at a zebra crossing in a simplified setting (see figure
\ref{fig:zebra_overview}). We consider one case where the pedestrian also
optimizes some of the reward of the vehicle (progress towards goal), i.e. the
pedestrian is partially cooperative, vs the scenario where the pedestrian does
not consider any of the vehicle's reward.
We implemented the iterative variant of the \ac{fancy} algorithm. We roll-out 12
time-steps and initialize the trajectories with each agent standing still. Please refer to figure \ref{fig:zebra} for the results. The
optimization takes 0.5s on a Titan Xp for a single trajectory and 0.6s for a
batch of 100 trajectories, opening up the possibility of probing multiple
initializations in real-time. For more details on the experiment (e.g. reward
setup), please refer to the supplementary material.

The \textit{resulting} behaviour varies significantly,
underscoring the importance of being able to formulate reward functions on the
continuum between full cooperation and no cooperation at all. In particular, the socially-minded pedestrian only considers a small part of the overall car specific reward function (namely, the goal reward). In a fully cooperative setup, both agents would be forced to share all of their rewards which complicates engineering a reward function that matches real-world behaviour.

\section{CONCLUSIONS}
We presented a novel algorithm for predicting boundedly rational human behaviour in multi-agent stochastic games efficiently. 
Furthermore, the algorithm can be used to infer the rewards of those agents.
We demonstrated its advantage for inferring rewards when agents execute their actions independently with limited communication. Also, we illustrated how diverse rewards affect the behaviour of agents and the variance inherent in
maximum-entropy methods that model boundedly rational agents.
We leave for future work the application to real-world traffic data.
\section{SUPPLEMENTARY MATERIAL}
\section*{ADDITIONAL BASELINES}
We provide two additional baselines. One that compares the policy roll-outs of iterative GS-CIOC to those of a policy obtained from value iteration. The other checks how well the single-agent CIOC algorithm \cite{Levine2012} performs on multi-agent data.
\subsection*{Value Iteration Baseline}
We will expand on the analysis in the experimental section of the main paper.
Namely, we introduce a baseline to understand if iterative GS-CIOC finds a
reasonable policy approximation. Figure \ref{fig:zebra_overview} summarizes
the toy example we consider. A pedestrian crosses the street, and an approaching
car stops so that the pedestrian can cross safely. 

The environment transitions are linear with $x_{kt} = x_{kt-1} + u_{kt}$. The reward setup is as follows. Both agents receive
\begin{itemize}
\item a quadratic lane keeping reward
\item a quadratic velocity reward 
\item a quadratic goal reward
\end{itemize}

Additionally, the car receives an interaction reward of the form 

\begin{multline}
\log(|x_c| + 0.1) * \textnormal{sigmoid}(-x_c + 6.) *
\textnormal{sigmoid}(x_c) \\ * \textnormal{sigmoid}(-y_p)
\end{multline}

where $x_c$ is the offset to the intersection point for the car in x-direction
and $y_p$ is the offset to the intersection point for the pedestrian in
y-direction. The pedestrian may also receive a reward for hurrying up while crossing. Namely, the pedestrian also receives the quadratic goal reward of the car on top of its own reward. We refer to that pedestrian as being socially minded.

\begin{figure}[h]
\includegraphics[width=0.47\textwidth]{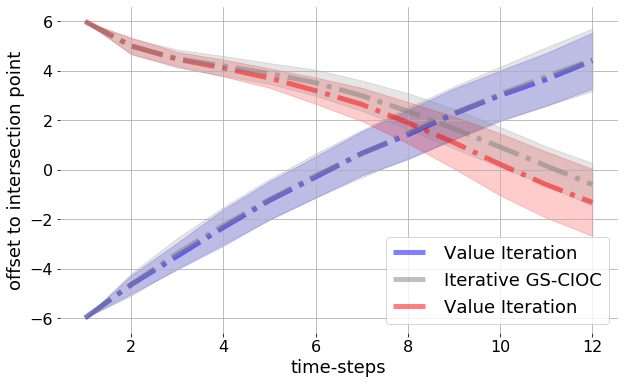}
\centering
\caption{The \textcolor{blue}{pedestrian} is not socially minded and crosses the zebra
crossing without paying much attention to the \textcolor{red}{car}. The blue and red curves
indicate a solution that we obtained from value iteration. The grey curves
correspond to the solution of iterative GS-CIOC. Due to the localized nature of
GS-CIOC, the mean trajectory does not match the solution of the value iteration
perfectly. Given that GS-CIOC will consider the reward function up to a
second-order Taylor expansion, the estimates of the standard deviation are not identical to those
of the value iteration solution. The pedestrian only experiences
quadratic rewards, while the car receives non-quadratic interaction
rewards.}\label{fig:asocial}
\end{figure}

\begin{figure}[h]
\includegraphics[width=0.47\textwidth]{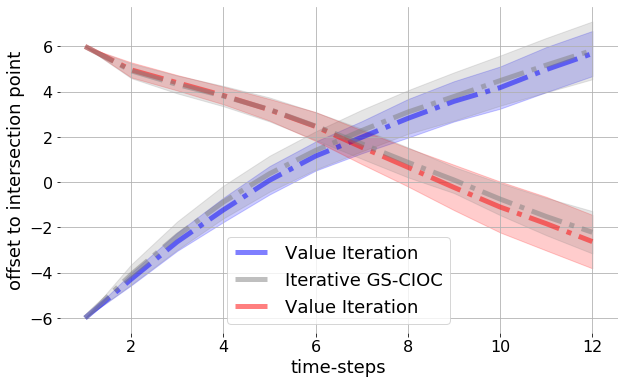}
\centering
\caption{Same as figure \ref{fig:asocial}, except here the \textcolor{blue}{pedestrian} values the
progress of the \textcolor{red}{car}, i.e. the pedestrian is more socially minded. The red and
blue curves indicate the solution obtained from value iteration and the grey
curves correspond to the solution of iterative GS-CIOC.} \label{fig:social}
\end{figure}

We apply two methods to obtain a policy. The first method is
iterative GS-CIOC, as described in the main paper. The second is value iteration
using the soft-Bellman equation and a discretization of the state and action
space. We perform value iteration for one agent at a time while the policy of the other agent is fixed, i.e. the policies are updated iteratively. These updates continue until the policies of both agents converge. We compare the policy roll-outs of both methods in figures
\ref{fig:asocial} and \ref{fig:social}.

Overall, the mean policy of iterative GS-CIOC converges close to the mean policy of the value iteration algorithm. Though, it gets stuck in local optima. Depending on the initial trajectories iterative GS-CIOC can slightly over- or under-shoot the value iteration reference trajectories. Figures \ref{fig:asocial} and \ref{fig:social} correspond to an initialization of both agents standing still at their respective starting positions (-6 and 6).

Apart from the mean policy roll-outs, the standard deviations show differences as well. Given only quadratic rewards, the iterative GS-CIOC standard deviation from the mean for the pedestrian (blue) in figure \ref{fig:asocial} agrees well with that of the value iteration algorithm. Though, the standard deviation is lower for the car (red) due to the non-quadratic nature of the interaction reward as GS-CIOC approximates the reward function with a second-order Taylor expansion. In other words, GS-CIOC may provide a biased estimate of the standard deviation.


\subsection*{Single-Agent CIOC Baseline}
Here we show which reward parameters CIOC infers given the ground-truth
roll-outs of GS-CIOC in a non-cooperative reward setup. Since CIOC assumes a
single agent setup, applying the algorithm to infer rewards for interacting agents
corresponds to the assumption that other agents are non-reacting dynamic
obstacles that follow pre-computed paths. We have already
established that GS-CIOC produces the same mean trajectories as a multi-agent
formulation of CIOC that we termed M-CIOC given cooperative rewards. The only
difference was the resulting variance which we were able to explain in terms of
the perfect coordination between agents for M-CIOC vs the decentralized
execution in GS-CIOC. 

The experimental setup is the same as that for the
quadratic reward experiments in the main paper. Given 2000 roll-outs of GS-CIOC we apply
CIOC to each agent. We repeat the procedure three times with varying seed
values for the random number generators. The ranges of the reward parameters
that we obtain are $\alpha = [0.245-0.248(0.40), 1.017-1.043(1.50),
1.945-1.954(2.50)]$ for one agent and $\alpha=[0.122-0.123(0.20),
0.712-0.723(1.00), 2.042-2.060(3.00)]$ for the other agent. The ground-truth
reward parameters are indicated inside the brackets.  CIOC cannot recover the
correct rewards. In particular, the difference cannot be explained with a single reward scaling factor. The scaling factors would be $[0.613-0.619, 0.678-0.696,
0.778-0.782]$ for one agent and $[0.612-0.617, 0.712-0.723, 0.681-0.687]$ for
the other agent. Therefore, the variation of the scaling factor is up to 20\% in
between the reward parameters of a single agent (e.g. 0.619 vs 0.778). This suggests that CIOC may
incur a significant bias when estimating the reward parameters of interacting
agents.
\section*{DERIVATION OF VALUE RECURSION FORMULAS}

This section provides additional details on the derivation of the value
recursion formulas in the methods section of the paper. Please read the methods
section first. Also, our calculations follow a similar path like those in the
supplementary material of \cite{Levine2012} for the single-agent case (LQR
Likelihood Derivation, section B,
\href{http://graphics.stanford.edu/projects/cioc/supplement.pdf}{Link}).
\subsection*{Linearization of the Dynamics}

\begin{align}
    \bar x_{kt} & = A_{kt} \bar x_{kt-1} + B_{kt} \bar
    u_{kt}\label{eq:linearization}\\
    A_{kt} & = \frac{\partial \bar x_{kt}}{\partial \bar x_{kt-1}}\\
    B_{kt} & = \frac{\partial \bar x_{kt}}{\partial \bar u_{kt-1}}
\end{align}

\subsection*{Relevant Integrals}
\begin{equation}
    \int (\bar u_{jt}^T\phi_u)\mathcal{N}(\mu_{jt};\Sigma_{jt})d\bar u_{jt} =
    \mu_{jt}^T\phi_u
    \label{eq:generalgaussint}
\end{equation}
\begin{equation}
    \int\Big(\bar x_{jt}^T \phi_x\Big)\mathcal{N}(\mu_{jt};\Sigma_{jt})d\bar
    u_{jt} = \chi_{jt}^T \phi_x
\end{equation}
\begin{multline}
    \int (\bar u_{jt}^T\mathbf{\Phi_u}\bar
    u_{jt})\mathcal{N}(\mu_{jt};\Sigma_{jt})d\bar u_{jt} = \\
    \mu_{jt}^T\mathbf{\Phi_u}\mu_{jt} + \sum_{nm}\Big(\mathbf{\Phi_u}\Big)_{nm}\Big(\Sigma_{jt}\Big)_{nm}
\end{multline}
\begin{multline}
    \int\Big(\bar x_{jt}^T \mathbf{\Phi_x}\bar x_{jt}\Big)\mathcal{N}(\mu_{jt};\Sigma_{jt})d\bar u_{jt} = \\
    \chi_{jt}^T \mathbf{\Phi_x}\chi_{jt}
     +\sum_{nm}\Big(B_{jt}^T\mathbf{\Phi_x}
     B_{jt}\Big)_{nm}\Big(\Sigma_{jt}\Big)_{nm}\label{eq:generalgaussintend}
\end{multline}
where
\begin{equation}
    \chi_{jt} = A_{jt}\bar x_{jt-1} + B_{jt}\mu_{jt}
\end{equation}
is the mean state that agent $j$ transitions to. 

\subsection*{Quadratic Reward Approximation}
In contrast to \cite{Levine2012}, we will also consider reward terms where states
and actions mix. The full second-order Taylor expansion of the reward is given by
\begin{multline}
    r(\bar x_{it}, \bar x_{jt}, \bar u_{it}, \bar u_{jt}) \approx \\r_t + 
    \begin{bmatrix}\bar x_{it}\\ \bar x_{jt}\\ \bar u_{it}\\ \bar u_{jt}\end{bmatrix}^T H_t \begin{bmatrix}\bar x_{it}\\ \bar x_{jt}\\ \bar u_{it}\\ \bar u_{jt}\end{bmatrix}
    +\begin{bmatrix}\bar x_{it}\\ \bar x_{jt}\\ \bar u_{it}\\ \bar u_{jt}\end{bmatrix}^T
    g_t\\
    = r_t + \bar x_{jt}^T\Big(H_{(x_jx_i)t}\bar x_{it}\Big) + \bar u^T_{jt}\Big(
    H_{(u_ju_i)t}\bar u_{it} \Big)\\+\bar u_{it}^T\Big(H_{(u_ix_j)t}\bar
    x_{jt}\Big) +\bar u_{jt}^T\Big(H_{(u_jx_i)t}\bar x_{it}\Big)
    \\+\sum_{k\in\{i,j\}} \Bigg[\bar
    x_{kt}^T\Big(\frac{1}{2}H_{(x_kx_k)t}\Big)\bar
    x_{kt} + \bar u^T_{kt} \Big(\frac{1}{2}H_{(u_ku_k)t}\Big)\bar u_{kt} \\+
    \bar u_{kt}^Tg_{u_kt} + \bar x_{kt}^Tg_{x_kt}\Bigg]
\end{multline}
\begin{align}
H_{(x_nx_m)t} &= \frac{\partial^2 r}{\partial \bar x_{nt} \partial \bar
x_{mt}}\\
H_{(u_nu_m)t} &= \frac{\partial^2 r}{\partial \bar u_{nt} \partial \bar
u_{mt}}\\
H_{(u_nx_m)t} &= \frac{\partial^2 r}{\partial \bar u_{nt} \partial \bar
x_{mt}}\\
g_{x_nt} & = \frac{\partial r}{\partial \bar x_{nt}}\\
g_{u_nt} & = \frac{\partial r}{\partial \bar u_{nt}}
\end{align}
We assume that the derivatives commute, i.e.
\begin{equation}
    H_{(u_m x_n)t}^T = H_{(x_n u_m)t}
\end{equation}
We can expand some of the $\bar x_{kt}$ using the linearization of the dynamics in
(\ref{eq:linearization}). This allows us to introduce a few convenient re-definitions of the
Hessian matrices and gradients which we will use for the remainder of the derivation.
\begin{align}
    \tilde H_{(ji)t} & = B_{jt}^T H^T_{(u_i x_j)t} + H_{(u_j x_i)t} B_{it} +
    H_{(u_j u_i)t}\\
    \tilde H_{(ii)t} & = \frac{1}{2} \Big(H_{(u_i x_i)t}B_{it} + B_{it}^TH_{(u_i
    x_i)t}^T\Big) + H_{(u_i u_i)t}\\
    \tilde H_{(jj)t} & = \frac{1}{2} \Big(H_{(u_j x_j)t} B_{jt} + B_{jt}^T H_{(u_j
    x_j)t}^T\Big) + H_{(u_j u_j)t}\\
    \hat H_{(nm)t} & = H_{(x_n x_m)t}\\
    \tilde g_{it} & = g_{u_it} + H_{(u_i x_i)t}A_{it} \bar x_{it-1} + H_{(u_i
    x_j)t}A_{jt} \bar x_{jt-1}\\
    \tilde g_{jt} & = g_{u_jt} + H_{(u_j x_j)t}A_{jt} \bar x_{jt-1} + H_{(u_j
    x_i)t}A_{it} \bar x_{it-1}\\
    \hat g_{nt} & = g_{x_nt}
\end{align}
The reward approximation given the re-defined Hessians and gradients is as follows.
\begin{multline}
    r(\bar x_{it}, \bar x_{jt}, \bar u_{it}, \bar u_{jt}) \approx \\ r_t + \bar
    x_{jt}^T\Big(\hat H_{(ji)t}\bar x_{it}\Big) + \bar u^T_{jt}\Big(\tilde
    H_{(ji)t}\bar u_{it} \Big)
    \\+\sum_{k\in\{i,j\}}^2 \Bigg[\bar x_{kt}^T\Big(\frac{1}{2}\hat H_{(kk)t}\Big)\bar
    x_{kt} + \bar u^T_{kt} \Big(\frac{1}{2}\tilde H_{(kk)t}\Big)\bar u_{kt} +
    \\\bar u_{kt}^T\tilde g_{kt} + \bar x_{kt}^T\hat g_{kt}\Bigg]
    \label{eq:rewardtaylor}
\end{multline}
As we have seen in the methods section, the value function will be quadratic in
the states given the reward approximation.
\begin{multline}
    V_{it+1}(\bar x_{1t}, \bar x_{2t})  =
    \bar x^T_{jt} \Big(\hat V_{(ji)t}\bar x_{it}\Big) +\\ \sum_{k=1}^2\left[\bar x^T_{kt} \Big(\frac{1}{2}\hat V_{(kk)t}\Big)\bar x_{kt} + \bar x^T_{kt}\hat v_{kt}\right] + c
    \label{eq:valuetaylor}
\end{multline}
\begin{align}
\hat V_{(nm)t} &= \frac{\partial^2 r}{\partial \bar x_{nt} \partial \bar
x_{mt}}\\
\hat v_{nt} & = \frac{\partial r}{\partial \bar x_{nt}}\\
\end{align}
The constant $c$ in the value function is going to be irrelevant
(does not affect policy) and will be dropped from now on.

\subsection*{Q-function}
\begin{multline}
    Q_{it}(\bar x_{it-1}, \bar x_{jt-1}, \bar u_{it}) = \\ \int p(\bar x_{it}, \bar x_{jt}|\bar x_{it-1}, \bar x_{jt-1}, \bar u_{it}, \bar u_{jt}) \pi_{jt}(\bar u_{jt}|\bar x_{it-1}, \bar x_{jt-1})\\ \Big(r(\bar x_{it}, \bar x_{jt}, \bar u_{it}, \bar u_{jt}) + V_{it+1}(\bar x_{it}, \bar x_{jt})\Big)d\bar u_{jt}d\bar x_{it}d\bar x_{jt}
    \label{eq:twoagentbellman}
\end{multline}
As we have argued in the methods section, the policy $\pi_{jt}$ is a Gaussian policy of the form
$\pi_{jt}\sim \exp\left(-\frac{1}{2}(\bar u_{jt} -
\mu_{jt})^T\Sigma_{jt}^{-1}(\bar u_{jt} - \mu_{jt})\right)$. Substituting
\ref{eq:rewardtaylor} and \ref{eq:valuetaylor} in \ref{eq:twoagentbellman}
results in quite a few terms of the type
\ref{eq:generalgaussint}-\ref{eq:generalgaussintend} and trivial
terms (simple Gaussian integral).

We collect the terms so that they resemble the integrals in
\ref{eq:generalgaussint}-\ref{eq:generalgaussintend} and define 
\begin{equation}
    \Phi_x = \frac{1}{2}\hat V_{(jj)t} + \frac{1}{2}\hat H_{(jj)t}
\end{equation}
\begin{equation}
    \phi_x = \hat H_{(ji)t}\bar x_{it} + \hat V_{(ji)t}\bar x_{it} + \hat g_{jt} + \hat v_{jt}
\end{equation}
for the terms where $\bar x_{jt}$ needs to be integrated over and
\begin{equation}
    \Phi_u = \frac{1}{2}\tilde H_{(jj)t}
\end{equation}
\begin{equation}
    \phi_u = \tilde H_{(ji)t}\bar u_{it} + \tilde g_{jt}
\end{equation}
for those with $\bar u_{jt}$ as the integration variable.

After integration and dropping a few irrelevant constants (no state or action
dependencies) we get the following Q-function.
\begin{multline}
    Q_{it}(\bar x_{1t-1}, \bar x_{2t-1}, \bar u_{it}) = g(\bar x_{it}, \bar u_{it}) + \chi_{jt}^T\Phi_x\chi_{jt} \\\mu^T_{jt} \Phi_u \mu_{jt} + \chi_{jt}^T\phi_x(\bar x_{it}) + \mu_{jt}^T \phi_u(\bar u_{it})
\end{multline}
where
\begin{multline}
    g(\bar x_{it}, \bar u_{it}) = \bar x_{it}^T\Big(\frac{1}{2}\hat H_{(ii)t}\Big)\bar x_{it} + \bar u^T_{it} \Big(\frac{1}{2}\tilde H_{(ii)t}\Big)\bar u_{it} + \\\bar u_{it}^T\tilde g_{it} + \bar x_{it}^T\hat g_{it} + \bar x^T_{it} \Big(\frac{1}{2}\hat V_{(ii)t}\Big)\bar x_{it} + \bar x^T_{it}\hat v_{it}
\end{multline}

\subsection*{Value Function}

To derive the value recursion for the decentralized setting, we
rearrange the values in the Q-function so that it resembles the
single-agent version. The integration $V_{it} = \log\int\exp(Q_{it}(\bar u_{it}))d\bar u_{it}$ that
provides us with the value function at time-step $t$ is then the same as that of
the single agent case. We give the single-agent result as a reference (taken from
supplementary material of
\cite{Levine2012}).

\hfill
\textcolor{blue}{\hrule}
\noindent\textcolor{blue}{\textbf{Single agent reference solution of
}\cite{Levine2012}.}
\begin{align}
Q_t(&\bar x_{t-1}, \bar u_t) = \frac{1}{2} \bar u_t^T \tilde H_t \bar u_t
+ \bar u_t^T \tilde g_t\\
&+\frac{1}{2}\underbrace{[A_t \bar x_{t-1} + B_t\bar u_t]}_{\bar
x_{t}}^T\underbrace{\Big[\hat H_t + \hat
V_t\Big]}_{\hat Q_t}[A_t\bar x_{t-1} + B_t \bar u_t]\\
&+[A_t \bar x_{t-1} + B_t\bar u_t]^T\underbrace{[\hat g_t + \hat v_t]}_{\hat
q_t}
\end{align}
The resulting value function $V_t = \log\int\exp(Q_t(\bar u_t))d\bar u_t$ is 
\begin{align}
V_t(\bar x_{t-1}) &= \frac{1}{2} \bar x_{t-1}^TA_t^T\hat Q_tA_t\bar
x_{t-1} + \bar x_{t-1}^TA_t^T\hat q_t\\
&-\frac{1}{2}\tilde \mu_t^T\tilde M_t\tilde \mu_t - \frac{1}{2}\log|-\tilde
M_t|
\end{align}
with
\begin{align}
\tilde 
\mu_t & = \tilde M_t^{-1}\Big(\tilde g_t + B_t^T\hat q_t +
B_t^T\hat Q_t A_t\bar
x_{t-1})\\
\tilde M_t & = \tilde H_t + B_t^T \hat Q_tB_t
\end{align}
\textcolor{blue}{\hrule}
\hfill

We use the single-agent solution as a template for the two-agent derivation of
the value function given the Q-function. 
\begin{multline}
    Q_{it}(\bar x_{1t-1}, \bar x_{2t-1}, \bar u_{it}) =\\ \frac{1}{2}\bar
    x_{it}^T\hat Q_{it} \bar x_{it} + \frac{1}{2}\bar u_{it}^T \tilde
    H_{(ii)t}\bar u_{it} + \bar u_{it}\tilde g_{ijt}^*
    + \bar x_{it}^T\hat q_{ijt} + C_{jt}
\end{multline}
with
\begin{equation}
    \hat Q_{it} = \hat H_{(ii)t} + \hat V_{(ii)t}
\end{equation}
\begin{equation}
    \tilde g_{ijt}^* = \tilde g_{it} + \tilde H_{(ij)t}\mu_{jt}
\end{equation}
\begin{equation}
    \hat q_{ijt} = \hat g_{it} + \hat v_{it} + \hat H_{(ij)t}\chi_{jt} + \hat V_{(ij)t}\chi_{jt}
\end{equation}
\begin{equation}
    C_{jt} = \chi_{jt}^T\Phi_x \chi_{jt} + \mu^T_{jt} \Phi_u \mu_{jt} + \chi_{jt}^T[\hat g_{jt} + \hat v_{jt}] + \mu_{jt}^T\tilde g_{jt}
\end{equation}
\begin{multline}
    C_{jt} = \frac{1}{2}\bar x_{jt-1}^TA^T_{jt}\hat Q_{jt}A_{jt}\bar x_{jt-1} +
    \frac{1}{2}\mu^T_{jt} \tilde M_{jt} \mu_{jt} + \\ \bar
    x_{jt-1}^TA^T_{jt}\hat Q_{jt}B_{jt}\mu_{jt} + \chi_{jt}^T[\hat g_{jt} +
    \hat v_{jt}] + \mu_{jt}^T\tilde g_{jt}
\end{multline}
The value function for timestep $t$ is
\begin{multline}
    V_{it}(\bar x_{it-1}, \bar x_{jt-1}) = \frac{1}{2}\bar x_{it-1}^TA^T_{it}\hat Q_{it}A_{it}\bar x_{it-1} +\\ \bar x_{it-1}^TA_{it}^T\hat q_{ijt} - \frac{1}{2}\mu_{it}^T\tilde M_{it}\mu_{it} - \frac{1}{2}\log |-\tilde M_{it}| + C_{jt}
\end{multline}
\begin{equation}
    \mu_{it} = -\tilde M^{-1}_{it}\Big(\tilde g_{ijt}^* + B^T_{it}\hat q_{ijt} + B^T_{it}\hat Q_{it}A_{it}\bar x_{it-1}\Big)
    \label{eq:mean}
\end{equation}
\begin{equation}
    \tilde M_{it} = \tilde H_{(ii)t} + B^T_{it}\hat Q_{it}B_{it}
\end{equation}
$\mu_{it}$ corresponds to the expected action of agent i considering the
interaction effects with agent j. $\tilde M_{it}$ is the precision matrix that
corresponds to the actions of agent i. Therefore, the solution
resembles the single agent case.
\subsection*{Deriving Mean}
We expand (\ref{eq:mean}) by substituting the definitions of $\tilde g^*$,
$\tilde g$, $\hat
q$ and $\chi$. The result is a linear equation in the mean actions $\mu_{mt}$.
\begin{align}
    -\boldsymbol{\mu}_{it} =& \tilde M_{it}^{-1}\Big[\tilde g_{it} + B_{it}^T\hat
    q_{it}\Big] + \tilde M_{it}^{-1}B_{it}^T\hat Q_{it}A_{it}\bar
    x_{it-1}\nonumber\\ + &\tilde M^{-1}_{it}B_{it}\hat Q_{(ij)t}A_{jt}\bar
    x_{jt-1} + \tilde M^{-1}_{it}\tilde M_{(ij)t}\boldsymbol{\mu}_{jt}\nonumber\\
    = & \tilde M_{it}^{-1}\Big(\alpha_i^{(i)} + \beta_i^{(i)}\bar x_{it-1} +
    \gamma_{ij}^{(i)}\bar x_{jt-1} + \delta_{ij}^{(i)}\boldsymbol{\mu}_{jt}\Big)\\
    \Rightarrow &\Big[\delta_{ij}^{(i)}\Big(\tilde M_{jt}^{-1}\Big)^{(j)}\delta_{ji}^{(j)} - \Big(\tilde M_{it}\Big)^{(i)}\Big]\mu_{it} =\nonumber\\ &\alpha_i^{(i)} - \delta_{ij}^{(i)}\Big(\tilde M_{jt}^{-1}\Big)^{(j)}\alpha_j^{(j)} + \nonumber\\ &\Big(\beta_i^{(i)} - \delta_{ij}^{(i)}\Big(\tilde M_{jt}^{-1}\Big)^{(j)}\gamma_{ji}^{(j)}\Big)\bar x_{it-1} +\nonumber\\
    &\Big(\gamma_{ij}^{(i)} - \delta_{ij}^{(i)}\Big(\tilde M_{jt}^{-1}\Big)^{(j)}\beta_{j}^{(j)}\Big)\bar x_{jt-1}
\end{align}
with
\begin{align}
    \alpha_i^{(i)} = & \Big[g_{u_it} + B_{it}^T\hat q_{it}\Big]\\
    \beta_i^{(i)} = & \hat Q_{it}^\dagger A_{it}\\
    \gamma_{ij}^{(i)} = & \hat Q_{(ij)t}^\dagger A_{jt}\\
    \delta_{ij}^{(i)} = & \tilde M_{(ij)t}\\
    \hat Q^\dagger_{(mn)t} =& H_{u_mx_n} + B^T_{mt}\hat Q_{(mn)t}
\end{align}
The $^{(i)}$ index indicates which agent the derivatives/ value functions refer
to. We indicate the index on the left side of the above equations but drop it otherwise
for brevity.

\subsection*{Value Recursion Formulas}

Now that we understand what the Q-function and the policy look like, we determine the set of recursive equations that provide us with the value function matrices at timestep $t$.
\begin{align}
    V_{it}(&\bar x_{it-1}, \bar x_{jt-1}) = \textcolor{blue}{\Bigg[}\frac{1}{2}\bar
    x_{it-1}^TA^T_{it}\hat Q_{it}A_{it}\bar x_{it-1} \nonumber\\&+\bar
    x_{it-1}^TA_{it}^T\hat q_{it} - \frac{1}{2}\bar\mu_{it}^{T}\tilde
    M_{it}\bar\mu_{it} - \frac{1}{2}\log |-\tilde
    M_{it}|\nonumber\\&- \frac{1}{2}\mu_{it}^{*T}\tilde
    M_{it}\mu_{it}^* - \bar\mu_{it}^{T}\tilde M_{it}\mu_{it}^*
\textcolor{blue}{\Bigg]_{\textnormal{single agent}}}\nonumber\\&
    +\textcolor{red}{\Bigg[}
    \frac{1}{2}\chi_{jt}^TD_{\chi\chi}\chi_{jt} +
    \frac{1}{2}\mu_{jt}^TD_{\mu\mu}\mu_{jt} + \mu_{jt}^TD_{\mu\chi}\chi_{jt}
    \nonumber\\&+ \mu_{jt}^TD_{\mu x}\bar x_{it-1} + \chi_{jt}^TD_{\chi x}\bar
    x_{it-1} + \mu_{jt}^TD_{\mu} \nonumber\\&
    + \mu_{jt}^TD^*_{\mu} + \chi_{jt}^TD_{\chi} + \chi_{jt}^TD^*_{\chi}\textcolor{red}{\Bigg]_{\textnormal{interaction}}}
\end{align}
\begin{align}
    \bar\mu_{it} =&-\tilde M^{-1}_{it}\Big(g_{u_it} + B^T_{it}\hat q_{it}
    + B^T_{it}\hat Q_{it}A_{it}\bar x_{it-1}\Big)\\
    \mu_{it}^{*}=&-\tilde M_{it}^{-1}\Big( H_{u_ix_i}A_ix_{it-1} +
    H_{u_ix_j}A_jx_{jt-1}\Big)
\end{align}
\begin{equation}
    D_{\chi\chi} = \hat Q_{(jj)t} - \hat Q_{(ji)t}B_{it}\tilde M_{it}^{-1}B_{it}^T\hat Q_{(ij)t}
\end{equation}
\begin{equation}
    D_{\mu\mu} = \tilde H_{(jj)t} - \tilde H_{(ji)t}\tilde M_{it}^{-1}\tilde H_{(ij)t}
\end{equation}
\begin{equation}
    D_{\mu\chi} = -\tilde H_{(ji)t}\tilde M_{it}^{-1}B^T_{it}\hat Q_{(ij)t}
\end{equation}
\begin{equation}
    D_{\mu x} = -\tilde H_{(ji)t}\tilde M^{-1}_{it}B_{it}^T\hat Q_{it}A_{it}
\end{equation}
\begin{equation}
    D_{\chi x} = \hat Q_{(ji)t}A_{it} - \hat Q_{(ji)t}B_{it}\tilde M^{-1}_{it} B^T_{it}\hat Q_{it}A_{it}
\end{equation}
\begin{equation}
    D_{\mu} = g_{u_jt} - \tilde H_{(ji)t}\tilde M^{-1}_{it}[g_{u_it} +
    B_{it}^T\hat q_{it} ]
\end{equation}
\begin{align}
    D^*_{\mu} &= H_{(u_jx_i)t}A_{it}\bar x_{it-1} + H_{(u_jx_j)t}A_{jt}\bar
    x_{jt-1}\nonumber\\&- \tilde H_{(ji)t}\tilde
    M^{-1}_{it}\Big[H_{(u_ix_i)t}A_{it}\bar x_{it-1} + H_{(u_ix_j)t}A_{jt}\bar
    x_{jt-1}\Big]
\end{align}
\begin{equation}
    D_{\chi} = \hat g_{jt} + \hat v_{jt} - \hat Q_{(ji)t}B_{it}\tilde M^{-1}_{it}[\tilde g_{it} + B^T_{it}\hat q_{it}]
\end{equation}
\begin{align}
    D^*_{\chi} &= - \hat Q_{(ji)t}B_{it}\tilde M^{-1}_{it}\Big[H_{(u_ix_i)t}A_{it}\bar
    x_{it-1} \nonumber\\&+ H_{(u_ix_j)t}A_{jt}\bar x_{jt-1}\Big]
\end{align}
$\mu_{it}^*$, $D_\mu^*$ and $D_\chi^*$ are correction terms due to the mixing of
states and actions in the reward function.

In the next step, we collect all the $\bar x_{kt-1}$ terms to reconstruct the
value function at $t$. In order to do so we need
to consider the state dependency of $\mu$ which did not play any role so far as
it depends only on the states at $t-1$ and not $t$.
\begin{equation}
    \mu_{jt} = \nu_{jt} + \Pi_{jt}\bar x_{it-1} + \Omega_{jt} \bar x_{jt-1}
\end{equation}

\begin{align}
    -\frac{1}{2}\mu_{it}^{*T}&\tilde M_{it}\mu_{it}^*
    -\bar\mu_{it}^{T}\tilde M_{it}\mu^*_{it}= \nonumber\\
    &-\textcolor{blue}{\bar x_{it-1}^T} A_{it}^T H_{x_i u_i} \tilde
    M_{it}^{-1}\Big(g_{u_{it}} + 
    B_{it}^T\hat q_{it}\Big)\nonumber\\ &- \textcolor{red}{\bar x_{jt-1}^T} A_{jt}^T
    H_{x_j u_i} 
    \tilde M_{it}^{-1}\Big(g_{u_{it}} + B_{it}^T \hat q_{it}\Big)\nonumber\\
    & - \textcolor{blue}{\bar x_{it-1}^T} A_{it}^T \hat Q_{it}B_{it}\tilde
    M_{it}^{-1} H_{u_i x_i}A_{it} \textcolor{blue}{\bar x_{it-1}} \nonumber\\
    &-\textcolor{red}{\bar x_{jt-1}^T} A_{jt}^TH_{x_ju_i}\tilde M_{it}^{-1}B_{it}^T\hat 
    Q_{it}A_{it}\textcolor{blue}{\bar x_{it-1}}\nonumber\\
    & -\textcolor{red}{\bar x_{jt-1}^T}A_{jt}^TH_{x_ju_i}\tilde
    M_{it}^{-1}H_{u_ix_i}A_{it}\textcolor{blue}{\bar x_{it-1}}\nonumber\\
    & -\frac{1}{2}\textcolor{red}{\bar x_{jt-1}^T}A_{jt}^TH_{x_ju_i}\tilde
    M_{it}^{-1}H_{u_ix_j}A_{jt}\textcolor{red}{\bar x_{jt-1}}\nonumber\\
    & -\frac{1}{2}\textcolor{blue}{\bar x_{it-1}^T}A_{it}^TH_{x_iu_i}\tilde
    M_{it}^{-1}H_{u_ix_i}A_{it}\textcolor{blue}{\bar x_{it-1}}
\end{align}
\begin{align}
    \frac{1}{2}\chi_{jt}^TD_{\chi\chi}\chi_{jt} = &\frac{1}{2}\textcolor{red}{\bar x_{jt-1}^T}(A_{jt} + B_{jt}\Omega_{jt})^TD_{\chi\chi}(A_{jt} + B_{jt}\Omega_{jt})\textcolor{red}{\bar x_{jt-1}}\nonumber\\&
    + \frac{1}{2}\textcolor{blue}{\bar x_{it-1}^T}\Pi_{jt}^TB^T_{jt}D_{\chi\chi}B_{jt}\Pi_{jt}\textcolor{blue}{\bar x_{it-1}}\nonumber\\
    & +\textcolor{red}{\bar x_{jt-1}^T}(A_{jt} + B_{jt}\Omega_{jt})^TD_{\chi\chi}B_{jt}\Pi_{jt}\textcolor{blue}{\bar x_{it-1}}\nonumber\\
    & +\textcolor{red}{\bar x_{jt-1}^T}(A_{jt} + B_{jt}\Omega_{jt})^TD_{\chi\chi}B_{jt}\nu_{jt}\nonumber\\
    & +\textcolor{blue}{\bar
    x_{it-1}^T}\Pi_{jt}^TB_{jt}^TD_{\chi\chi}B_{jt}\nu_{jt} + const
\end{align}
\begin{align}
    \frac{1}{2}\mu_{jt}^TD_{\mu\mu}\mu_{jt} = &\frac{1}{2}\textcolor{red}{\bar x_{jt-1}^T}\Omega_{jt}^TD_{\mu\mu}\Omega_{jt}\textcolor{red}{\bar x_{jt-1}}\nonumber\\&
    + \frac{1}{2}\textcolor{blue}{\bar x_{it-1}^T}\Pi_{jt}^TD_{\mu\mu}\Pi_{jt}\textcolor{blue}{\bar x_{it-1}}\nonumber\\
    & +\textcolor{red}{\bar x_{jt-1}^T}\Omega_{jt}^TD_{\mu\mu}\Pi_{jt}\textcolor{blue}{\bar x_{it-1}}\nonumber\\
    & +\textcolor{red}{\bar x_{jt-1}^T}\Omega_{jt}^TD_{\mu\mu}\nu_{jt}\nonumber\\
    & +\textcolor{blue}{\bar x_{it-1}^T}\Pi_{jt}^TD_{\mu\mu}\nu_{jt} + const
\end{align}
\begin{align}
    \mu_{jt}^TD_{\mu\chi}\chi_{jt} &= \Big[\textcolor{red}{\bar x_{jt-1}^T} A_{jt}^TD_{\mu\chi}^T\nu_{jt}\nonumber\\
    &+ \textcolor{red}{\bar x_{jt-1}^T} A_{jt}^TD_{\mu\chi}^T\Pi_{jt}\textcolor{blue}{\bar x_{it-1}}\nonumber\\
    &+ \textcolor{red}{\bar x_{jt-1}^T} A_{jt}^TD_{\mu\chi}^T\Omega_{jt}\textcolor{red}{\bar x_{jt-1}}\Big]\nonumber\\
    &+ \Big[\textcolor{red}{\bar x_{jt-1}^T}\Omega_{jt}^T(D_{\mu\chi}B_{jt})\Omega_{jt}\textcolor{red}{\bar x_{jt-1}}\nonumber\\&
    + \textcolor{blue}{\bar x_{it-1}^T}\Pi_{jt}^T(D_{\mu\chi}B_{jt})\Pi_{jt}\textcolor{blue}{\bar x_{it-1}}\nonumber\\
    & +\textcolor{red}{\bar x_{jt-1}^T}\Omega_{jt}^T(D_{\mu\chi}B_{jt} + B_{jt}^TD_{\mu\chi}^T)\Pi_{jt}\textcolor{blue}{\bar x_{it-1}}\nonumber\\
    & +\textcolor{red}{\bar x_{jt-1}^T}\Omega_{jt}^T(D_{\mu\chi}B_{jt} + B_{jt}^TD_{\mu\chi}^T)\nu_{jt}\nonumber\\
    & +\textcolor{blue}{\bar x_{it-1}^T}\Pi_{jt}^T(D_{\mu\chi}B_{jt} +
    B_{jt}^TD_{\mu\chi}^T)\nu_{jt}\Big] + const
\end{align}
\begin{align}
    \mu_{jt}^TD_{\mu x}\bar x_{it-1} =& \textcolor{blue}{\bar x_{it-1}^T}D_{\mu x}^T \nu_{jt} + \textcolor{blue}{\bar x_{it-1}^T}D_{\mu x}^T\Pi_{jt}\textcolor{blue}{\bar x_{it-1}}\nonumber\\
    & + \textcolor{red}{\bar x_{jt-1}^T}\Omega_{jt}^TD_{\mu x}\textcolor{blue}{\bar x_{it-1}}\\
    \chi_{jt}^TD_{\chi x}\bar x_{it-1} &= \textcolor{red}{\bar x_{jt-1}^T}A_{jt}^TD_{\chi x}\textcolor{blue}{\bar x_{it-1}} \nonumber\\
    & + \textcolor{blue}{\bar x_{it-1}^T}(D_{\chi x}^TB_{jt}) \nu_{jt} \nonumber\\&+\textcolor{blue}{\bar x_{it-1}^T}(D_{\chi x}^TB_{jt})\Pi_{jt}\textcolor{blue}{\bar x_{it-1}}\nonumber\\
    & + \textcolor{red}{\bar x_{jt-1}^T}\Omega_{jt}^T(B_{jt}^TD_{\chi x})\textcolor{blue}{\bar x_{it-1}}\\
    \mu_{jt}^TD_\mu = & \textcolor{blue}{\bar x_{it-1}^T}\Pi_{jt}^TD_\mu +
    \textcolor{red}{\bar x_{jt-1}^T}\Omega_{jt}^TD_\mu + const
\end{align}
\begin{align}
    \chi_{jt}^TD_\chi = & \textcolor{red}{\bar x_{jt-1}^T}A_{jt}^TD_\chi\nonumber\\
    & + \textcolor{blue}{\bar x_{it-1}^T}\Pi_{jt}^T(B_{jt}^TD_\chi) + \textcolor{red}{\bar x_{jt-1}^T}\Omega_{jt}^T(B_{jt}^TD_\chi) \nonumber\\&+ const
\end{align}
\begin{align}
    \mu_{jt}^T D^*_{\mu} &=
    \textcolor{blue}{\bar x_{it-1}^T} A_{it}^T\Big(H_{x_i u_j} - H_{x_i u_i} \tilde
    M_{it}^{-1}\tilde H_{(ij)t}\Big)\nu_{jt}\nonumber\\
    &+ \textcolor{red}{\bar x_{jt-1}^T }
    A_{jt}^T\Big(H_{x_j u_j} - H_{x_j u_i} 
    \tilde M_{it}^{-1}\tilde 
    H_{(ij)t}\Big)\nu_{jt}\nonumber\\
    &+ \textcolor{blue}{\bar x_{it-1}^T}\Pi_{jt}^T \Big(H_{u_j 
    x_i} - \tilde H_{(ji)t}
    \tilde M_{it}^{-1}H_{u_i x_i} \Big)A_{it}\textcolor{blue}{\bar x_{it-1}}\nonumber\\
    &+\textcolor{blue}{\bar x_{it-1}^T}\Pi_{jt}^T \Big(H_{u_j 
    x_j} - \tilde H_{(ji)t}
    \tilde M_{it}^{-1}H_{u_i x_j}
    \Big)A_{jt}\textcolor{red}{\bar x_{jt-1}}\nonumber\\
    &+\textcolor{red}{\bar x_{jt-1}^T}\Omega_{jt}^T \Big(H_{u_j 
    x_j} - \tilde H_{(ji)t}
    \tilde M_{it}^{-1}H_{u_i x_j} \Big)A_{jt}\textcolor{red}{\bar x_{jt-1}}\nonumber\\
    &+\textcolor{red}{\bar x_{jt-1}}^T\Omega_{jt}^T \Big(H_{u_j 
    x_i} - \tilde H_{(ji)t}
    \tilde M_{it}^{-1}H_{u_i x_i} \Big)A_{it}\textcolor{blue}{\bar x_{it-1}}\nonumber\\
\end{align}

\begin{align}
    \chi_{jt}^T &D^*_{\chi} =-\textcolor{blue}{\bar x_{it-1}^T} A_{it}^TH_{x_iu_i}\tilde
    M_{it}^{-1}B_{it}^T\hat Q_{(ij)t}B_{jt}\nu_{jt}\nonumber\\ &-
    \textcolor{red}{\bar x_{jt-1}^T} A_{jt}^TH_{x_ju_i}\tilde
    M_{it}^{-1}B_{it}^T\hat Q_{(ij)t}B_{jt}\nu_{jt}\nonumber\\ &-
    \textcolor{blue}{\bar x_{it-1}^T}\Pi_{jt}^T B_{jt}^T\hat
    Q_{(ji)t}B_{it}\tilde M_{it}^{-1}H_{u_ix_i}A_{it}\textcolor{blue}{\bar
    x_{it-1}}\nonumber\\ &-\textcolor{blue}{\bar x_{it-1}^T}\Pi_{jt}^TB_{jt}^T
    \hat Q_{(ji)t}B_{it}\tilde M_{it}^{-1}H_{u_ix_j}A_{jt}\textcolor{red}{\bar
    x_{jt-1}}\nonumber\\ &-\textcolor{red}{\bar
    x_{jt-1}^T}\Big(A_{jt}+B_{jt}\Omega_{jt}\Big)^T \hat Q_{(ji)t}B_{it}\tilde
    M_{it}^{-1}H_{u_ix_j}A_{jt}\textcolor{red}{\bar x_{jt-1}}\nonumber\\
    &-\textcolor{red}{\bar x_{jt-1}^T}\Big(A_{jt} + B_{jt}\Omega_{jt}\Big)^T
    \hat Q_{(ji)t}B_{it}\tilde M_{it}^{-1}H_{u_ix_i}A_{it}\textcolor{blue}{\bar
    x_{it-1}}\nonumber\\
\end{align}

Colours indicate which agent the states belong to.
Now, we collect the terms and also make use of the single agent solution
provided by \cite{Levine2012} for reward functions where states and actions do
not mix.

First, we collect the terms that resemble the following expression.
\begin{equation}
    \frac{1}{2}\bar x_{nt-1}^T\hat V_{(nn)t-1} \bar x_{nt-1}
\end{equation}
\begin{align}
    \hat V_{(ii)t-1} =& \Pi_{jt}^T \Big[\tilde M_{jt} - \tilde
    M_{(ji)t}\tilde M_{it}^{-1}\tilde M_{(ij)t}]\Pi_{jt} +\nonumber\\
    &\Big[A_{it}^T\hat Q^*_{(ij)t} - A_{it}^T\hat Q_{it}^*\tilde
    M_{it}^{-1}\tilde M_{(ij)t}\Big]\Pi_{jt}+\nonumber\\
    &\Pi_{jt}^T\Big[A_{it}^T\hat Q^*_{(ij)t} - A_{it}^T\hat
    Q_{it}^*\tilde M_{it}^{-1}\tilde M_{(ij)t}\Big]^T\nonumber\\
    &\textcolor{blue}{+ A^T_{it}\hat Q_{it}A_{it}}
    \textcolor{blue}{-A_{it}^T\hat Q_{it}^*\tilde M_{it}^{-1}\hat
    Q_{it}^\dagger A_{it}}
\end{align}
With
\begin{equation}
    \tilde M_{(nm)t} = B_{nt}^T\hat Q_{(nm)t}B_{mt}+\tilde H_{(nm)t}
\end{equation}
\begin{equation}
    \hat Q^*_{(nm)t} = H_{x_nu_m} + \hat Q_{(nm)t}B_{mt}
\end{equation}
\begin{equation}
    \hat Q^\dagger_{(mn)t} = H_{u_mx_n} + B^T_{mt}\hat Q_{(mn)t}
\end{equation}
Blue indicates the single agent solution.
\begin{align}
    \hat V_{(jj)t-1} = & \Omega^T_{jt}\Big[\tilde M_{jt} - \tilde M_{(ji)t}\tilde
    M_{it}^{-1}\tilde M_{(ij)t}\Big]\Omega_{jt} + \nonumber\\
    & \Big[A^T_{jt}\hat Q^*_{(jj)t} - A^T_{jt}\hat Q^*_{(ji)t}\tilde
    M_{it}^{-1}\tilde M_{(ij)t}\Big]\Omega_{jt} + \nonumber\\
    & \Omega_{jt}^T\Big[A^T_{jt}\hat Q^*_{(jj)t} - A^T_{jt}\hat
    Q^*_{(ji)t}\tilde M_{it}^{-1}\tilde M_{(ij)t}\Big]^T +\nonumber\\
    & A_{jt}^T\hat Q_{(jj)t}A_{jt} - A_{jt}^T \hat Q^*_{(ji)t}\tilde
    M_{it}^{-1}\hat Q^\dagger_{(ij)t}A_{jt}
\end{align}
The
structure of the equation closely mirrors that of $\hat V_{(ii)t}$.

\noindent Next, we collect the following terms.
\begin{equation}
    \bar x_{nt-1}^T\hat V_{(nm)t-1} \bar x_{mt-1}
\end{equation}
\begin{align}
    \hat V_{(ji)t-1} = & \Omega_{jt}^T\Big[\tilde M_{jt} - \tilde M_{(ji)t}\tilde
    M_{it}^{-1}\tilde M_{(ij)t}\Big]\Pi_{jt} + \nonumber\\
    & \Omega_{jt}^T\Big[A_{it}^T\hat Q^*_{(ij)t} - A_{it}^T\hat
    Q^*_{it}\tilde M_{it}^{-1}\tilde M_{(ij)t}\Big]^T + \nonumber\\
    & \Big[A_{jt}^T\hat Q^*_{(jj)t} - A_{jt}^T\hat Q^*_{(ji)t}\tilde
    M_{it}^{-1}\tilde M_{(ij)t}\Big]\Pi_{jt} + \nonumber\\
    & A_{jt}^T\hat Q_{(ji)t}A_{it} - A_{jt}^T\hat Q^*_{(ji)t}\tilde M_{it}^{-1}\hat Q^\dagger_{(ii)t}A_{it}
\end{align}
And finally, we collect the terms that resemble the following expression.
\begin{equation}
    \bar x_{nt-1}^T\hat v_{nt-1}
\end{equation}
\begin{align}
    \hat v_{jt-1} &= \Omega_{jt}^T\Big[\tilde M_{jt} - \tilde M_{(ji)t}\tilde
    M_{it}^{-1}\tilde M_{(ij)t}\Big]\nu_{jt}\nonumber\\ & +\Big[A^T_{jt}\hat
    Q^*_{(jj)t} - A^T_{jt}\hat Q^*_{(ji)t}\tilde M_{it}^{-1}\tilde
    M_{(ij)t}\Big]\nu_{jt}\nonumber\\ & + \Omega_{jt}^T\Big(\tilde g_{jt} +
    B_{jt}^T\hat q_{jt} - \tilde M_{(ji)t}\tilde M_{it}^{-1}\Big[\tilde g_{it} +
    B_{it}^T\hat q_{it}\Big]\Big)\nonumber\\ & + A_{jt}^T\hat q_{jt} -
    A_{jt}^T\hat Q^*_{(ji)t}\tilde M^{-1}_{it}[\tilde g_{it} + B^T_{it}\hat
    q_{it}]
\end{align}
\begin{align}
    \hat v_{it-1} &= \Pi_{jt}^T\Big[\tilde M_{jt} - \tilde M_{(ji)t}\tilde
    M_{it}^{-1}\tilde M_{(ij)t}\Big]\nu_{jt} \nonumber\\ & +\Big[A^T_{it}\hat
    Q^*_{(ij)t} - A^T_{it}\hat Q^*_{(ii)t}\tilde M_{it}^{-1}\tilde
    M_{(ij)t}\Big]\nu_{jt} \nonumber\\ & +\Pi_{jt}^T\Big(\tilde g_{jt} +
    B_{jt}^T\hat q_{jt} - \tilde M_{(ji)t}\tilde M_{it}^{-1}\Big[\tilde g_{it} +
    B_{it}^T\hat q_{it}\Big]\Big)\nonumber\\ & \textcolor{blue}{+ A^T_{it}\hat
    q_{it}}\textcolor{blue}{ -A^T_{it}\hat Q^*_{(ii)t}\tilde
    M^{-1}_{it}\Big[\tilde g_{it} + B^T_{it}\hat q_{it}\Big]}
\end{align}
Again, blue indicates the single agent solution.

\section*{COMMENT ON EXTENSION TO N AGENTS}
We may define an N-agent extension of the soft-Bellman equation that reflects the two-agent case discussed in the main paper.
\begin{multline}
    Q_i(x_{t-1}, u_{it}) = \\
    \int p(x_t|x_{t-1}, u_{1t}, ..., u_{Nt})\Pi_{k=1}^N\pi_{k}(u_{kt}|x_{t-1})
    \\\Big(r_i(x_{t},
    u_{1t}, ..., u_{Nt}) \\+ \gamma \log \int \exp(Q_i(x_{t}, u_{it+1}))
    du_{it+1}\Big)du_{jt}dx_t \label{eq:nagent}
\end{multline}
The environment transitions $p(x_t|x_{t-1}, u_{1t}, ..., u_{Nt})$ will be deterministic again. 
Approximating the reward function by a second-order Taylor expansion in the states and actions 
\begin{multline}
    r(\bar x_{1t}, ..., \bar x_{Nt}, \bar u_{1t}, ..., \bar u_{Nt}) \approx \\r_t + 
    \begin{bmatrix}\bar x_{1t}\\ ...\\ \bar x_{Nt}\\ \bar u_{1t}\\...\\ \bar u_{Nt}\end{bmatrix}^T H_t \begin{bmatrix}\bar x_{1t}\\ ...\\ \bar x_{Nt}\\ \bar u_{1t}\\...\\ \bar u_{Nt}\end{bmatrix}
    +\begin{bmatrix}\bar x_{1t}\\ ...\\ \bar x_{Nt}\\ \bar u_{1t}\\...\\ \bar u_{Nt}\end{bmatrix}^T
    g_t
\end{multline}
will make the integral in (\ref{eq:nagent}) tractable. Again, the policies of the other agents will reduce to Gaussian distributions.

As with the two-agent case, the resulting Q-function will be quadratic in the actions $\bar u_{it}$. Thus, we will be able to determine the value function analytically as well given the integral $V_{it} = \log\int\exp(Q_{it}(\bar u_{it}))d\bar u_{it}$. In other words, the overall derivation will resemble that of the two-agent case. Though new value recursion matrices and equations will appear, namely, $\hat V_{nmt}$ matrices that describe the interaction of other agents. Given multiple counterparts agent i also needs to consider how other agents react to each other, increasing the complexity of the value recursion formulas. We leave the derivation and empirical verification of the N-agent case for future work.

\bibliographystyle{icml2020}
\bibliography{root.bib}

\end{document}